\pdfoutput=1
\documentclass[11pt]{article}

\usepackage{ACL2023}

\usepackage{times}
\usepackage{latexsym}

\usepackage[T1]{fontenc}

\usepackage[utf8]{inputenc}

\usepackage{microtype}

\usepackage{inconsolata}

\usepackage{amsmath}
\usepackage{adjustbox}
\usepackage{multirow}
\usepackage{subcaption}
\usepackage{caption}
\usepackage{balance} 
\usepackage{graphicx}
\usepackage{makecell}
\usepackage{amssymb}
\usepackage{anyfontsize}
\usepackage{arydshln}
\usepackage{bm}
\usepackage[section]{placeins}
\usepackage{float}
\usepackage{cleveref}
\usepackage{booktabs}
\usepackage{xcolor}
\usepackage{listings}
\usepackage{algorithm} 
\usepackage{algorithmic}
\usepackage{framed}
\usepackage{multicol}
\usepackage{mdframed}
\definecolor{shadecolor}{rgb}{0.92,0.92,0.92}

\newcommand{\methodname}{\textsc{Teller}}

\title{\methodname: A Trustworthy Framework For Explainable, Generalizable and Controllable Fake News Detection}

\author{Hui Liu\textsuperscript{1,2} \quad \quad  \quad
  Wenya Wang\textsuperscript{2} \quad \quad \quad   Haoru Li\textsuperscript{3} \quad \quad  \quad Haoliang Li\textsuperscript{1}\\
  \textsuperscript{1}City University of Hong Kong \\
    \textsuperscript{2}Nanyang Technological University\\
  \textsuperscript{3}University of Electronic Science and Technology of China\\ 
\text{liuhui3-c@my.cityu.edu.hk}, \text{wangwy@ntu.edu.sg},  \text{lihaoru114@gmail.com},\text{haoliang.li@cityu.edu.hk}
}

\begin{document}
\maketitle
\begin{abstract}
The proliferation of fake news has emerged as a severe societal problem, raising significant interest from industry and academia. While existing deep-learning based methods have made progress in detecting fake news accurately, their reliability may be compromised caused by the non-transparent reasoning processes, poor generalization abilities and inherent risks of integration with large language models (LLMs). To address this challenge, we propose {\methodname}, a novel framework for trustworthy fake news detection that prioritizes explainability, generalizability and controllability of models. This is achieved via a dual-system framework that integrates cognition and decision systems, adhering to the principles above. The cognition system harnesses human expertise to generate logical predicates, which guide LLMs in generating human-readable logic atoms. Meanwhile, the decision system deduces generalizable logic rules to aggregate these atoms, enabling the identification of the truthfulness of the input news across diverse domains and enhancing transparency in the decision-making process. 
Finally, we present comprehensive evaluation results on four datasets, demonstrating the feasibility and trustworthiness of our proposed framework. 
Our implementation is available at this link\footnote{\url{https://github.com/less-and-less-bugs/Trust_TELLER}}.
\end{abstract}

\section{Introduction}
Fake news has emerged as a prominent social problem due to the rampant dissemination facilitated by social media platforms \citep{zhou2020survey}. Additionally, the swift progress of generative artificial intelligence has further amplified this issue \citep{rochasurvey23}. While human fact-checking experts can accurately verify the authenticity of news, their efforts cannot scale with the overwhelming volume of online information. Consequently, researchers have turned to automatic fake news detection techniques.

Despite the improved predictive accuracy achieved by current deep learning-based detection approaches \citep{mj2019, qpeefmm2021,mehta2022tackling}, these methods suffer from the lack of transparency because of the black-box nature of neural networks \citep{shu2019defend} and a limited ability to generalize to unseen data of which the distribution is different from training data, given the inherent diversity of online information (e.g., topics, styles and media platforms) \citep{huitifs}. Moreover, the increasing integration with LLMs is prone to uncontrollable risks due to hallucinations and societal applications. Thus, a growing awareness emphasizes trustworthiness\footnote{In AI, trustworthiness refers to the extent to which an AI system can be trusted to operate ethically, responsibly, and reliably \citep{jobin2019global}. } of these systems \citep{liu-etal-2023-interpretable, sheng2022characterizing}.

\begin{figure}[t]
\centering
\includegraphics[width=1\linewidth]{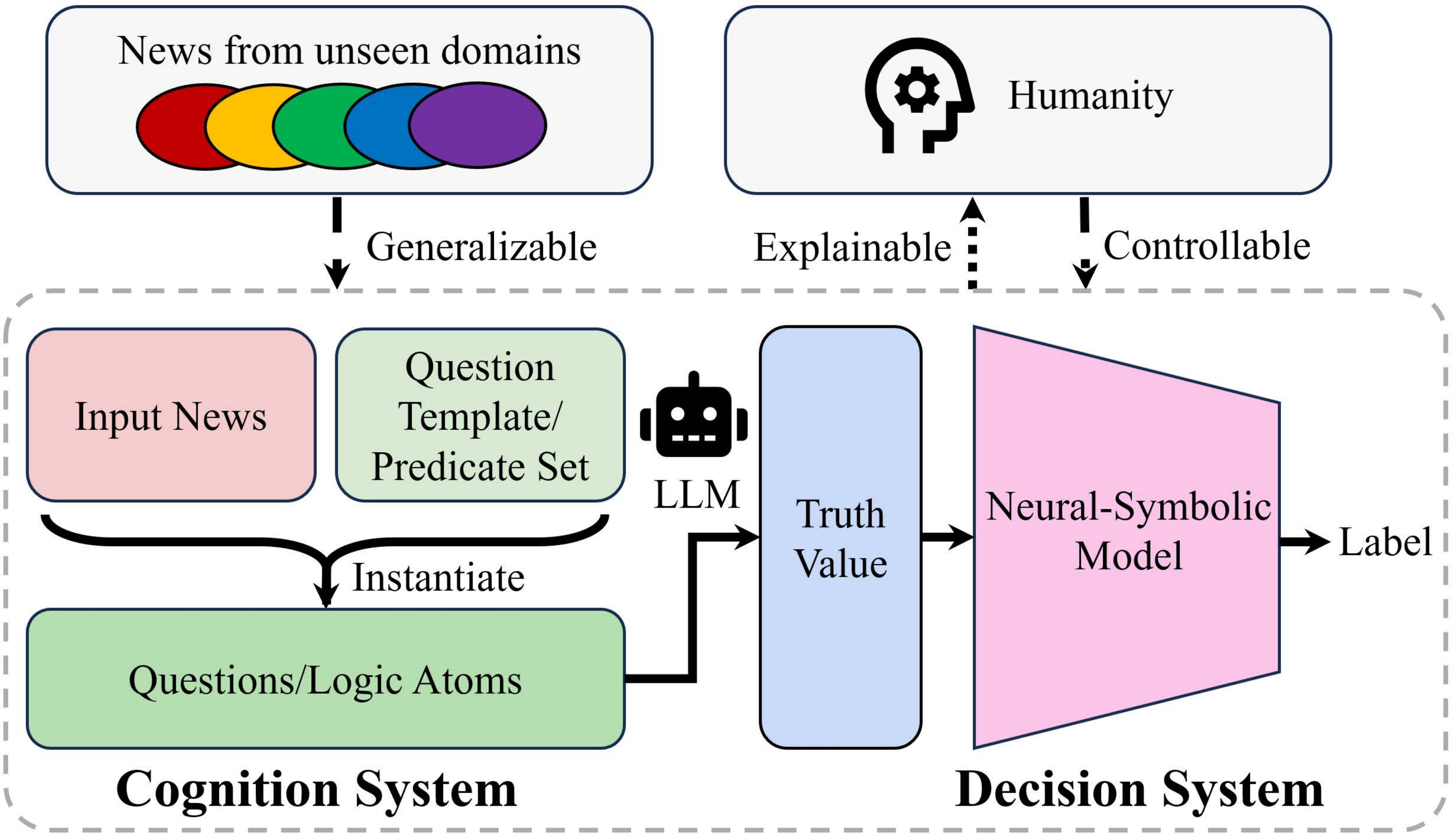}
\caption{Three crucial aspects of trustworthy fake news detection algorithms and the correlation between these principles and our dual-sytem framework \methodname.}
\label{figfirst}
\vspace{-20pt}
\end{figure}
Unfortunately, the characteristics of a trustworthy fake news detector remain an open question. Hence, based on recent surveys of Trustworthy AI \citep{li2023trustworthy, jobin2019global} and fake news detection \citep{shu2023survey}, we identify three crucial aspects that go beyond accuracy for fake news detection technologies: explainability, generalizability, and controllability. These aspects work collectively to enhance system security and trustworthiness.

Firstly, explainability refers to understanding how an AI model performs decision \citep{DBLP:journals/ai/Miller19}. This mechanism serves as a fundamental requirement for establishing end-user trust in these tools, as it enables the disclosure of complex reasoning processes and the identification of potential flaws in neural networks. Secondly, generalizability represents the capability to acquire knowledge from limited training data to predict accurately in unseen situations \citep{wang2023trustworthy}. Given the impracticality of exhaustively collecting and annotating vast amounts of data across various news domains, generalization ensures the affordable and sustainable deployment of data-driven fake news detection algorithms. Lastly, controllability encompasses the capacity for human guidance and  intervention in the behavior of models \citep{AIalignment}. This objective benefits models in understanding specific misinformation regulatory policies and rectifying deviations if necessary. While recent practices may satisfy the requirements of explainability \citep{eviwww22, liu-etal-2023-interpretable} or generalization \citep{rubostmultitask,mldgfakenews}, they often fail to adhere to all three principles simultaneously.

To this end, we propose {\methodname}, a \textbf{T}rustworthy framework for \textbf{E}xplainable, genera\textbf{L}izable and contro\textbf{L}labe d\textbf{E}tecto\textbf{R}, drawing inspiration from the dual-system theory\footnote{System 1 provides tools for intuitive, imprecise, and unconscious decisions akin to deep learning, while system 2 handles complex situations requiring logical and rational thinking akin to symbolic learning \citep{booch2021thinking}.} \citep{daniel2017thinking}. This framework abstracts the existing pipeline of fake news detection into two components: the cognition and decision systems. As depicted in Fig.~\ref{figfirst}, the cognition system serves as the first step and is responsible for transforming meaningful human expertise from renowned journalism teams \citep{HKBU, Politifact} into a set of Yes/No question templates that correspond to logic predicates. These decomposed questions are then answered using LLMs, which provide truth values for corresponding logic atoms. 

On the other hand, the decision system, empowered by a differentiable neural-symbolic model \citep{pix2rule},  
can integrate the output of the cognition system to deduce the final authenticity of input news by leveraging domain invariant logic rules learned from data automatically. This visible logic-based ensemble can mitigate the negative effects caused by inaccurate predictions of LLMs and allow for the correction of unreasonable rules through adjusting the weights in the model manually to align with human expertise.

Our framework ensures explainability by incorporating human-readable question templates (predicates) and a transparent decision-making process based on logic rules. This interpretability further enables the flexibility to adjust rules and enhances the model's robustness against false LLM predictions, thereby guaranteeing controllability. Moreover, our model exhibits generalizability, attributed to the generalizable performance of LLMs combined with reliable human experience as guidance and the utilization of the neural-symbolic model, which can learn domain-generalizable rules.

To summarize, the contributions of this work include: 1) We introduce a systematic framework comprising cognition and decision modules, aiming to uphold three crucial principles for establishing a trustworthy fake news detection system: explainability, generalizability, and controllability. 2) We validate the effectiveness of our framework by conducting comprehensive experiments using various LLMs on four benchmarks. The results demonstrate the feasibility and trustworthiness of {\methodname} across different scenarios.

\section{Related Work}
\subsection{Trustworthy AI}
Establishing comprehensive trustworthiness in AI is non-trivial due to its multi-objective nature, including robustness, security, transparency, fairness, safety, and ethical standards \citep{jobin2019global}. Achieving such trustworthiness necessitates considering the entire lifecycle of an AI system, spanning from data preparation and algorithm design, development, and deployment to management and governance \citep{li2023trustworthy, eykholt2018robust}. Recent researchers have explored diverse approaches to enhance AI trustworthiness across various goals and stages to address this challenge. 
For example, regarding algorithm design, several topics, such as transfer learning, federated learning, and interpretable AI, have been proposed to improve models' robustness, security, and transparency. Moreover, the deployment of AI systems necessitates external government oversight, particularly for AGI \citep{bengiomanaging}. Although our work focuses on enhancing the trustworthiness of detection systems from the algorithm design aspect, we acknowledge that there is still much room for improvement to achieve the ultimate goal.

\subsection{Trustworthy Fake News Detection}
Recent fake news detection research has witnessed a notable paradigm shift from prioritizing accuracy to considering trustworthiness. In line with our work, we primarily examine studies that aim to enhance algorithms' explainability, generalizability, and controllability. 

Regarding explainability, \citet{shu2019defend, eviwww22, evikdd23} suggested obtaining key evidence for interpretation based on feature importance, while \citet{liu-etal-2023-interpretable} utilized logic clauses to illustrate the reasoning processing. However, these methods still need to be more transparent due to their probabilistic nature and complex architecture. Furthermore, another group of works \citep{gptfakenews1, badactor,yue2024evidence, qi2024sniffer}, explored large generative language models (e.g., ChatGPT) and regarded the intermediate chain of thoughts as an explanation. Nevertheless, these explanations may not be reliable due to the hallucination phenomenon \citep{hallucination} and the misalignment problem of AGI \citep{AIalignment}. Moving on to generalizability, most methods, such as \citep{mldgfakenews, sqtkde, qpeefmm2021}, enhanced fake news detectors through transfer learning algorithms to learn domain-invariant features or domain-adaptive features. However, these methods inevitably introduce external costs of domain alignment, such as annotating domain labels. As for controllability, although some works \citep{silva2021embracing, mendes-etal-2023-human} incorporated the human-in-loop technique in data sampling and model evaluation, few works explore how to intervene and edit models to align with human expertise. More comparative discussion between {\methodname} and existing work can be found in Appendix \ref{sec:appendixf}. 

\section{Methodology}
\begin{figure*}[t]
\centering
\includegraphics[width=0.9\linewidth]{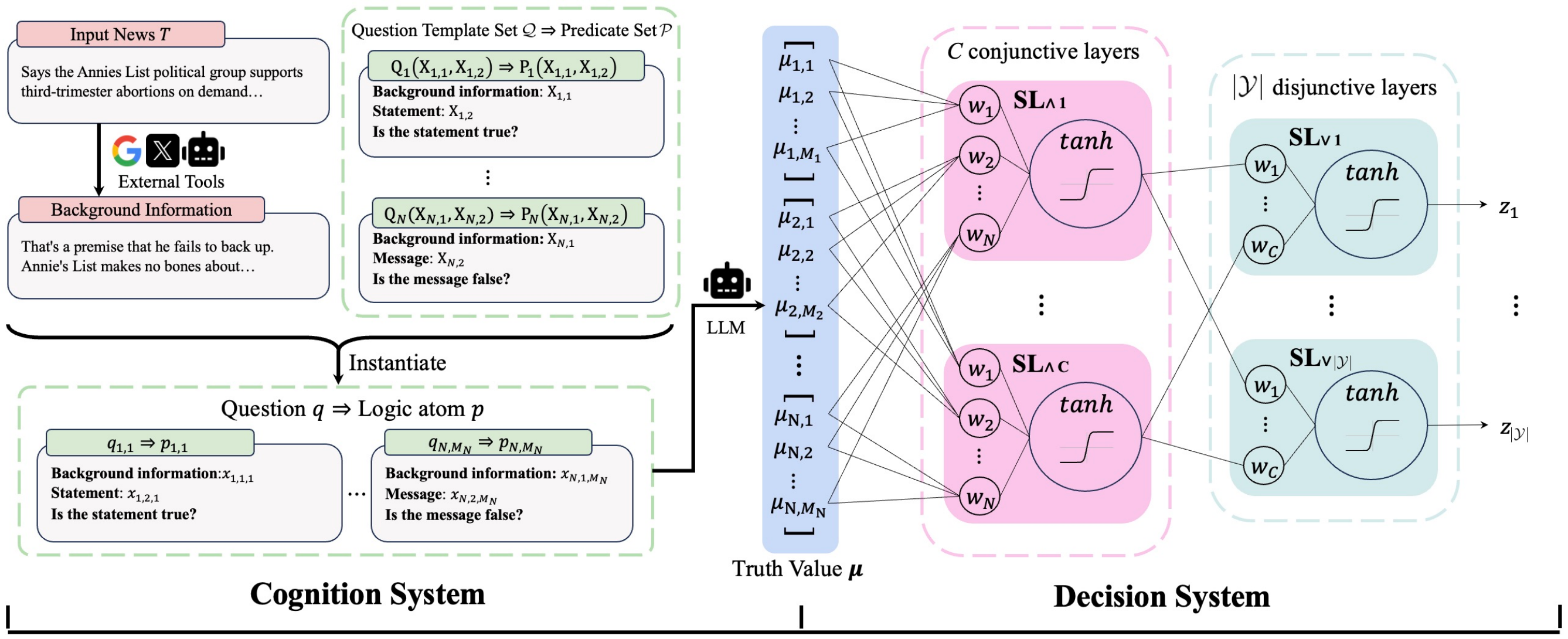}
\caption{The architecture of the proposed framework {\methodname}. $N$ represents the number of question templates (logic predicates), $M_i$ denotes the number of logic atoms corresponding to the $i$th predicate, $\mathcal{Y}$ denotes the truthfulness label set. The semantics of question templates and logic predicates are described in Table \ref{tab:logic-table}.}
\label{figsecond}
\vspace{-18pt}
\end{figure*}
Formally, given a piece of news $T$, the objective of the fake news detection task is to predict its label of truthfulness $y \in \mathcal{Y}$ where $\mathcal{Y}$ can fit in different levels of classification granularity. For example, in binary classification setting, $\mathcal{Y} = \{\text{true}, \text{false}\}$, and $T$ is identified as real (fake) when $y$ is true (false).

As depicted in Fig. \ref{figsecond}, {\methodname} involves two main components: cognition and decision systems. The cognition system decomposes human expertise into Yes/No question templates corresponding to logic predicates. When presented with a new input $T$, the templates and predicates can be instantiated to form questions and logic atoms. By leveraging the parametric knowledge inside LLMs and gathering additional information from external tools (e.g., search engines), the cognition system can generate answers to these questions, represented as truth values of logic atoms. Then, the decision system takes these truth values as input and generates interpretable logic clauses to debunk misinformation by a neural-symbolic model,  which can learn generic logic rules from data in an end-to-end manner.

\subsection{Cognition System}
To combat misleading information, existing deep learning-based algorithms fall short in gaining public trust, while fact-checking experts rigorously follow designated guidance and principles to facilitate transparent and fair evaluation. Our cognitive system aims to integrate the strengths of deep learning-based methods that can handle large-scale online information while maintaining the trustworthiness of manual checking.

\subsubsection{Predicate Construction}
To begin with, we describe the following symbol convention for clarity: calligraphic font $\mathcal{Q}$ and $\mathcal{P}$ for sets of question templates and predicates, capitalized letters $\mathrm{Q}$, $\mathrm{P}$, $\mathrm{X}$   for question templates, predicates, and variables, and corresponding lowercase letters  $q$, $p$, $x$ for instances of these entities (questions, logic atoms, values). The truth values of logic atoms are denoted by $\mu$.

Inspired by the well-established fact-checking process in Table \ref{tab:fact-check-process}, we initially decompose it into a question template set, denoted as $\mathcal{Q}$, containing eight questions as detailed in Appendix \ref{sec:appendixqt}.
Each template $\mathrm{Q}_i$ in $\mathcal{Q}$ consists of $N_i$ variables and can be transformed into an $N_i$-ary logic predicate ${\mathrm{P}_i(\mathrm{X}_{i, 1}, \ldots, \mathrm{X}_{i, N_i})}$ in $\mathcal{P}$. The logic semantics of $\mathrm{P}_i$ is interpreted as the affirmative answer to $\mathrm{Q}_i$ and its truth value $\mu_{i}$ represents the probability that $\mathrm{P}_i$ holds. For instance, take $\mathrm{Q}_1$ (i.e., "Background Information: $\mathrm{X}_{1,1}$. Statement: $\mathrm{X}_{1,2}$. Is the statement true?") in Fig. \ref{figsecond} as an example. The corresponding predicate $\mathrm{P}_1(\mathrm{X}_{1,1}, \mathrm{X}_{1,2})$ can be explained as "Given the background information $\mathrm{X}_{1,1}$, the statement $\mathrm{X}_{1,2}$ is true". 

For each predicate ${\mathrm{P}_i(\mathrm{X}_{i, 1}, \ldots, \mathrm{X}_{i, N_i})}$, we can instantiate the variables $\mathrm{X}_{i, 1}, \ldots, \mathrm{X}_{i, N_i}$ with the actual contents taken from any input news to obtain logic atoms. Since an input piece of news may contain multiple background information and statements (instantiations), we use $k$ to denote the $k$th instantiation where $1\leq k\leq\prod\limits_{{j=1}}^{N_i}|\mathrm{X}_{i,j}|$. Here $|\mathrm{X}_{i,j}|$ indicates the total number of possible instantiations for variable $\mathrm{X}_{i,j}$. Then we denote by $p_{i,k}$ the instantiated logic atom corresponding to the question $q_{i,k}$. Next, we introduce how to acquire the truth value of each logic atom.

\subsubsection{Logic evaluation with LLMs}
\label{sec:logicevaluationllm}
While decomposed questions can provide a comprehensive explanation of how the decision is made \citep{generalquestionjifan, specialquestionfan}, directly answering these questions poses a challenge due to the impracticality of annotating enormous data to train multiple models for different questions. To address this issue, we resort to the more general-purpose LLMs (e.g., FLAN-T5 \citep{flant5}, Llama2 \citep{llama2}, and GPT-3.5) as the foundation for effectively answering these questions. Existing LLMs can be categorized into two groups: $\text{LLM}_{open}$, such as FLAN-T5 and Llama2, where the logits of output vocabulary can be obtained, and $\text{LLM}_{close}$, such as GPT-3.5, where the logits are not accessible. 

To ensure compatibility with both categories of LLMs, we propose two strategies to obtain the final truth values of logic atoms. Concretely, we first input the question $q_{i,k}$ with a suffix (i.e., "Yes or No? Response:") to LLMs in order to measure their preference for the affirmative answer "Yes" versus the negative one "No".  This preference is subsequently used to compute the truth value of the corresponding logic atom $p_{i,k}$.

For $\text{LLM}_{open}$, we follow \cite{yesnoq, burns2022discovering} to obtain pre-softmax logits of  "Yes" and "No" tokens, denoted as $v_{Yes}$ and $v_{No}$ respectively. Compared with post-softmax logits, pre-softmax logits can mitigate the influence of other tokens in output vocabulary, particularly when LLMs tend to generate irrelevant tokens that may result in $v_{Yes}$ or $v_{No}$ becoming zero. Then the truth value $\mu$ for the logic atom $p$ (here we omit the underscript $i,k$ for ease of illustration) can be obtained as follows: 
\begin{equation}
\label{eq1}
\mu = 2\frac{e^{v_{Yes}}}{e^{v_{No}}+e^{v_{Yes}}}-1.
\end{equation}

For $\text{LLM}_{close}$, we sample $m$ times during decoding and count the frequency of "Yes" and "No" responses as $m_{Yes}$ and $m_{No}$. Then we compute
\begin{equation}
\label{eq2}
\mu = 2\frac{m_{Yes}}{m_{No}+m_{Yes}}-1.
\end{equation}
In either case, $\mu$ is in the range of $[-1, 1]$. When $\mu \in [-1,0)$, $\mu \in (0,1]$, and $\mu = 0$, the corresponding logic atom $p$ is evaluated as false, true, and unknown, respectively. Once the truth values of all logic atoms for a single predicate $\mathrm{P}_i$ (corresponding to a single question template) are obtained, we concatenate them as one vector, denoted as $\bm{\mu}_i$. Then we concatenate the value vectors for all predicates as the input for the final decision system. 

In conclusion, our cognition system can generate diversified questions and logic atoms based on the input news $T$.
These human-readable entities enhance explainability by showcasing potential intermediate reasoning steps and ensure controllability by allowing adjustments to $\mathcal{Q}$ and $\mathcal{P}$. Moreover, combining human expertise and LLMs provides the basis for the cognition system's satisfactory generalization performance in unseen domains.

\subsection{Decision System}
After acquiring responses to all questions, it is imperative to develop a decision system to effectively aggregate them to predict the label of the input news $T$ while preserving trustworthiness in the reasoning process. However, prevalent heuristic strategies (e.g., majority voting) lack the flexibility to handle complex relationships among different questions and cannot tolerate false predictions, and deep-learning-based models cannot be comprehended literally by humans \citep{wang2022interpretability}.


Hence, we utilize a neural-symbolic model, named Disjunctive Normal Form (DNF) Layer \citep{pix2rule, pixelrulplus}, as our decision system. This model includes conjunctive layers ($\mathrm{SL}_\land$) and disjunctive layers ($\mathrm{SL}_\lor$), which can progressively converge to symbolic semantics such as conjunction $\land$ and disjunction $\lor$ respectively during model training. Consequently, this model can automatically learn logic rules from data in an end-to-end manner, capturing generalizable relationships between logic predicates and the target label.  
As illustrated in Fig. \ref{figsecond}, we stack $C$ conjunctive layers $\mathrm{SL}_\land$ beneath $|\mathcal{Y}|$ disjunctive layers $\mathrm{SL}_\lor$ to construct the DNF Layer, where each $\mathrm{SL}_\lor$ corresponds to a truthfulness label $y\in \mathcal{Y}$.  

However, the original DNF Layer proposed in \citep{pix2rule} is not directly applicable to our work due to two issues.  Firstly, the truth value of logic atoms $\mu$ ranges in $[-1,1]$, while the original model can only handle values of $-1$ and $1$. Secondly, each logic atom in the original DNF Layer is treated differently
which loses logic semantics where atoms for the same logic predicate should share similar functionality. To address the aforementioned challenges, we propose a modified DNF layer which takes continuous values $\mu\in[-1,1]$ as input and assigns the same weight for those atoms instantiated from the same logic predicate. The detailed description of our modified DNF layer can be found in Appendix \ref{sec:appendixe}.

More concretely, in our proposed DNF Layer, every $\mathrm{SL}_\land$ takes truth values $\bm{\mu}$ of all logic atoms obtained in the cognition system as input, aiming to learn a conjunctive clause $\text{conj}=\bigwedge_{p_{i,k}\in \mathcal{A}} p_{i,k}$ where $\mathcal{A}\subseteq\{p_{1,1},\ldots,p_{N,M_N}\}$, referring to a subset of the complete logic atoms, and outputs the truth value of this conjunctive clause. Subsequently, each $\mathrm{SL}_\lor$ receives the truth values of $C$ conjunctive clauses to represent a disjunction of these conjunctions: $\bigvee_{c\in \mathcal{C}}\text{conj}_c$ where $\mathcal{C} \subseteq\{1,\ldots,C\}$, referring to a subset of all $\text{conj}$s. It then outputs the truth value of this disjunction formula, corresponding to the final probability that the input news $T$ is identified as the label $y$. Hence, each label $y$ will be associated with a DNF clause learned by the DNF layer. Intuitively, the conjunction simulates the idea that if the input news $T$ gives affirmative answers to some questions simultaneously, it is highly probable that it should be assigned to label $y$. On the other hand, the disjunction provides more flexibility by considering different alternatives (the output is true if at least one of the conj is true) which makes the final decision less sensitive to incorrect atom values due to wrong predictions given by LLMs. For example, assume the learned rules are $\text{conj}_1 \vee \text{conj}_2$ where $\text{conj}_1 = p_{1,1} \wedge p_{1,2}$ and $\text{conj}_2 = p_{2,1} \wedge p_{3,1}$. Suppose $\text{conj}_1$ is true, then we can conclude that $\text{conj}_1 \vee \text{conj}_2$ is true even if $\text{conj}_2$ gives an incorrect value. 

Last but not the least, we apply softmax function to the output of all disjunction layers $\mathrm{SL}_\lor$ to obtain the probability $\bm{z}\in \mathbb{R}^{|\mathcal{Y}|}$ for all possible labels. The entire decision system can be trained in an end-to-end fashion by minimizing the cross-entropy loss function as below: 
\begin{equation}
\label{eq:cross_entropy_loss}
\mathcal{L} = -\sum_{l=1}^{|\mathcal{Y}|}\mathbb{I}(y_l=y_T) \log z_{l},
\end{equation}
where $y_T$ represents the ground truth label of $T$. During inference, we select the label corresponding to the highest value in $\bm{z}$ as the final result.

In summary, our decision system can extract interpretable symbolic rules from data that exhibit robustness across diverse domains and enable intervention by adjusting weights in the DNF Layer to align with prior knowledge (refer to Appendix \ref{sec:appendixc}).


\section{Experiments}
In this section, we present the experiment setup and demonstrate the feasibility, explainability,  generalizability and controllability of {\methodname} through extensive experiments. 

\subsection{Experimental Setting}
\noindent\textbf{Dataset.} We conducted experiments using four challenging datasets, namely LIAR \citep{Liar}, Constraint \citep{Constraint}, PolitiFact, and GossipCop \citep{Fakenewsnet}. LIAR  comprises the binary classification and multi-classification setting with six fine-grained labels for truthfulness ratings. Moreover, \citet{Liar, liarwithevidence} curated relevant evidence (e.g., background information), serving as gold knowledge in an open setting. Constraint, PolitiFact and GossipCop are binary classification datasets related to COVID-19, politics, and entertainment domains, respectively. 

\noindent\textbf{LLMs.} We select the open-source FLAN-T5 and Llama2 series, which encompass various parameter sizes, as large language models for constructing the cognition system. We also conduct experiments using GPT-3.5-turbo on the LIAR dataset to examine the versatility of our framework.

\noindent\textbf{Baselines.} We compare our model against \textit{Direct}, \textit{Few-shot Direct}, \textit{Zero-shot COT},  \textit{Few-shot COT}, \textit{Few-shot Logic}. The baselines suffixed with \textit{Direct} involve prompting large language models (LLMs) to predict the label of input news directly; those suffixed with \textit{COT} utilize chain-of-thought techniques to enhance the performance of LLMs; those suffixed with \textit{Logic} replace the thought process in COT with questions paired with their answers. We exclusively implement COT-related methods using GPT-3.5-turbo because they show no improvement over \textit{Direct} on FLAN-T5 and  Llama 2, as shown in Table \ref{differentprompts}. Additionally, we compare with small models, including BERT and RoBERTa, analyzed in Appendix \ref{sec:appendixf}.

\noindent\textbf{Implementation Detail.} We evaluate the performance of our framework using the accuracy and Macro-F1, which accommodates class imbalance.  For each dataset, we train our decision system using the training split; select the optimal model based on its performance on the validation split; and report the results on the test split. To assess the generalizability of our model, we consider each dataset as a separate domain and train our models using the train split from source domains; choose the best model on the validation split of source ones; and report results on the test split from the target domain. Moreover, to highlight the robustness of our framework, we keep all hyperparameters fixed in each setting. Details of the experiment setting, data leakage analysis, baselines, and model training are elaborated in Appendix \ref{sec:appendixb}.


\begin{table*}[h]
\begin{adjustbox}{max width=\linewidth, center}
\begin{tabular}{cc|cc|cc|cc|cc}
\hline
\multirow{3}{*}{Large   Language Models} & \multirow{3}{*}{Method} & \multicolumn{4}{c}{Binary Classification} & \multicolumn{4}{c}{Multi-Classification} \\
\cline{3-10}
 &  & \multicolumn{2}{c}{Closed} & \multicolumn{2}{c}{Open} & \multicolumn{2}{c}{Closed} & \multicolumn{2}{c}{Open} \\
 \cline{3-10}
 &  & Acc(\%) & Macro-F1(\%) & Acc(\%) & Macro-F1(\%) & Acc(\%) & Macro-F1(\%) & Acc(\%) & Macro-F1(\%) \\
  \hline
FLAN-T5-small (80M) & Direct & $44.99$ & $31.63$ & $45.08$ & $32.41$ & $18.17$ & $9.28$ & $19.51$ & $10.13$ \\
 \hline
FLAN-T5-base (250M) & Direct & $54.02$ & $50.79$ & $61.47$ & $61.43$ & $19.43$ & $11.79$ & $21.40$ & $21.40$ \\
 \hline
\multirow{3}{*}{FLAN-T5-large (780M)} & Direct & $57.30$ & $52.20$ & $74.38$ & $73.84$ & $19.43$ & $17.84$ & $29.50$ & $24.95$ \\
\cdashline{2-10}
 & \methodname & $66.83_{(9.53\uparrow)}$ & $\mathbf{66.33}_{(14.13\uparrow)}$ & $77.76_{(3.38\uparrow)}$ & $77.32_{(3.49\uparrow)}$ & $\underline{\mathbf{26.99}}_{(7.55\uparrow)}$ & $18.04_{(0.20\uparrow)}$ & $33.67_{(4.17\uparrow)}$ & $27.50_{(2.55\uparrow)}$ \\
 & w/ Intervention & \colorbox{green!30}{$65.64$} & \colorbox{green!30}{$65.12$} & \colorbox{green!30}{$77.46$} & \colorbox{green!30}{$77.14$} &\colorbox{green!30}{ $26.28$} & $18.49$ & $35.25$ & $30.05$ \\
  \hline
\multirow{3}{*}{FLAN-T5-xl (3B)} & Direct & $58.89$ & $58.62$ & $75.97$ & $75.67$ & $19.67$ & $16.57$ & $29.43$ & $24.74$ \\
\cdashline{2-10}
 & \methodname & $62.36_{(3.48\uparrow)}$ & $60.18_{(1.56\uparrow)}$ & $78.75_{(2.78\uparrow)}$ & $78.55_{(2.88\uparrow)}$ & $24.31_{(4.64\uparrow)}$ & $17.40_{(0.83\uparrow)}$ & $33.52_{(4.09\uparrow)}$ & $27.22_{(2.48\uparrow)}$ \\
 & w/ Intervention & $63.65$ & $61.82$ & $79.34$ & $79.07$ & $25.57$ & $19.62$ & $34.46$ & $33.59$ \\
  \hline
\multirow{3}{*}{FLAN-T5-xxl (11B)} & Direct & $56.41$ & $56.08$ & $75.17$ & $75.15$ & $22.42$ & $18.31$ & $32.18$ & $28.12$ \\
\cdashline{2-10}
 & \methodname & $66.63_{(10.23\uparrow)}$ & $65.91_{(9.82\uparrow)}$ & $80.24_{(5.06\uparrow)}$ & $79.85_{(4.70\uparrow)}$ & $26.83_{(4.41\uparrow)}$ & $19.68_{(1.36\uparrow)}$ & $35.48_{(3.30\uparrow)}$ & $30.42_{(2.30\uparrow)}$ \\
 & w/ Intervention & $\mathbf{67.03}$ & $66.19$ & $80.73$ & $80.41$ & $\mathbf{26.91}$ & $\mathbf{21.30}$ & $35.88$ & $31.63$ \\
  \hline
\multirow{3}{*}{Llama2 (7B)} & Direct & $59.88$ & $59.19$ & $72.29$ & $69.63$ & $18.02$ & $9.97$ & $11.01$ & $6.88$ \\
\cdashline{2-10}
 & \methodname & $62.46_{(2.58\uparrow)}$ & $62.45_{(3.26\uparrow)}$ & $79.94_{(7.65\uparrow)}$ & $79.80_{(10.16\uparrow)}$ & $23.29_{(5.27\uparrow)}$ & $15.51_{(5.55\uparrow)}$ & $32.73_{(21.72\uparrow)}$ & $25.55_{(18.67\uparrow)}$ \\
 & w/ Intervention & $64.15$ & $62.77$ & $81.93$ & $81.84$ & $23.92$ & \colorbox{green!30}{$15.14$} & $34.30$ & $27.58$ \\
\hline
\multirow{3}{*}{Llama2 (13B)} & Direct & $56.90$ & $56.90$ & $69.31$ & $63.77$ & $7.32$ & $2.85$ & $10.86$ & $8.25$ \\
\cdashline{2-10}
 & Ours & $66.04_{(9.14\uparrow)}$ & $66.03_{(9.13\uparrow)}$ &$\mathbf{82.52}_{(13.21\uparrow)}$ & $\mathbf{82.37}_{(18.60\uparrow)}$ & $25.81_{(18.49\uparrow)}$ & $17.71_{(14.86\uparrow)}$ & $38.08_{(27.22\uparrow)}$ & $29.27_{(21.02\uparrow)}$ \\
 & w/ Intervention & $\underline{\mathbf{67.73}}$ & $\underline{\mathbf{66.97}}$ & $\underline{\mathbf{84.21}}$ & $\underline{\mathbf{84.03}}$ & \colorbox{green!30}{$25.10$} & \colorbox{green!30}{$16.78$} & $38.63$ & $30.60$ \\
  \hline
\multirow{6}{*}{GPT-3.5-turbo} & Direct & $42.40$ & $51.48$ & $76.27$ & $74.21$ & $20.46$ & $20.34$ & $26.20$ & $25.12$ \\
 & \methodname & - & - & $79.15_{(2.88\uparrow)}$ & $78.90_{(4.69\uparrow)}$ & - & - & $31.94_{(5.74\uparrow)}$ & $29.53_{(4.41\uparrow)}$ \\
 & Zero-shot COT & $30.88$ & $41.87$ & $72.49$ & $70.83$ & $7.16$ & $9.20$ & $39.81$ & $\mathbf{36.49}$ \\
 & Few-shot & $61.67$ & $64.05$ & $81.02$ & $81.00$ & $25.65$ & $\underline{\mathbf{25.56}}$ & $\underline{\mathbf{46.81}}$ & $\underline{\mathbf{44.61}}$ \\
 & Few-shot COT & $52.04$ & $56.15$ & $74.48$ & $76.21$ & $20.69$ & $17.20$ & $\mathbf{45.63}$ & $36.36$ \\
 & Few-shot Logic & $49.26$ & $48.85$ & $61.67$ & $60.92$ & $16.37$ & $13.98$ & $20.54$ & $19.22$\\
 \hline
\end{tabular}
\end{adjustbox}
\caption{Results on LIAR dataset. "Closed" represents the cognitive system does not have access to any external knowledge source, while "Open" indicates that it can utilize gold evidence collected by human experts. The best results for each setting are highlighted with bold numbers and an underline, whereas sub-optimal results are only highlighted in bold. The\colorbox{green!30}{number} indicates that the performance of \textit{w/ Intervention} is worse than \methodname. The number with $\uparrow$ indicates the performance gain of \methodname\ over \textit{Direct}.}
\vspace{-8pt}
\label{liarresults}
\end{table*}

\subsection{Feasibility Study}
To validate the feasibility of our framework, we compare it against multiple baselines across a wide range of LLMs and scenarios (e.g., different classification granularities) in Table \ref{liarresults} and Table \ref{mainresultsthree}. These results uncover two crucial findings listed below:
 
Firstly, our framework demonstrates satisfactory performance in fake news detection tasks. Specifically, in the binary classification setting, {\methodname} achieves an accuracy of approximately 76\% on the GossipCop dataset and over 80\% on the other three datasets. Notably, when utilizing Llama 2 (13B) to drive the cognition system, {\methodname} outperforms all GPT-3.5-turbo based methods by a significant margin. These results highlight the effectiveness of {\methodname} in distinguishing between fake and genuine news. In the multi-classification setting on the LIAR dataset, our framework consistently outperforms \textit{Direct} for FLAN-T5 and Llama2 series, even though these models may struggle to discriminate fine-grained labels. This observation underscores the capability of our decision system to mitigate the negative influences of noisy predictions in the cognition system, effectively unleashing the potential of LLMs through logic-based aggregation of answers to decomposed questions.

Secondly, our framework exhibits significant potential for the future. In the binary classification setting across four datasets, {\methodname} consistently outperforms \textit{Direct} in terms of accuracy and macro-F1 scores by an average of 7\% and 6\%, respectively. Considering the swift improvement of LLM intelligence, these results imply that the performance of our framework is likely to scale with the evolution of LLMs. Additionally, due to the notable performance difference between closed and open settings on the LIAR dataset, it is promising to integrate external tools to acquire extensive evidence from credible sources, such as official government websites, to enhance the performance of our systems.


\begin{table*}[h]
\begin{adjustbox}{max width=0.8\linewidth, center}
\begin{tabular}{cc|cc|cc|cc}
 \hline
\multirow{2}{*}{LLMs} & \multirow{2}{*}{Method} & \multicolumn{2}{c}{Constraint} & \multicolumn{2}{c}{PolitiFact} & \multicolumn{2}{c}{GossipCop} \\
\cline{3-8}
 &  & Acc(\%) & Macro-F1(\%) & Acc(\%) & Macro-F1(\%) & Acc(\%) & Macro-F1(\%) \\
 \hline
\multirow{3}{*}{FLAN-T5-large} & Direct & $78.06$ & $77.97$ & $56.62$ & $54.84$ & $67.43$ & $58.76$ \\
\cdashline{2-8}
 & {\methodname}& $80.32_{(2.27\uparrow)}$ & $80.11_{(2.14\uparrow)}$ & $67.65_{(11.03\uparrow)}$ & $67.65_{(12.81\uparrow)}$ & $69.53_{(2.10\uparrow)}$ & $59.39_{(0.63\uparrow)}$ \\
 & w/ Intervention & $80.46$ & $80.31$ & $68.38$ & $68.29$ & $70.28$ & $60.74$ \\
  \hline
\multirow{3}{*}{FLAN-T5-xl} & Direct & $75.32$ & $74.79$ & $55.88$ & $50.72$ & $67.73$ & $52.80$ \\
\cdashline{2-8}
 & \methodname & $83.77_{(8.45\uparrow)}$ & $83.66_{(8.88\uparrow)}$ & $68.82_{(9.14\uparrow)}$ & $64.68_{(13.95\uparrow)}$ & $69.58_{(1.85\uparrow)}$ & $58.72_{(5.91\uparrow)}$ \\
 & w/ Intervention & $83.95$ & $83.88$ & $69.12$ & $68.79$ & $72.23$ & $63.84$ \\
  \hline
\multirow{3}{*}{FLAN-T5-xxl} & Direct & $74.80$ & $73.23$ & $52.21$ & $43.65$ & $68.93$ & $52.82$ \\
\cdashline{2-8}
 & \methodname & $83.39_{(8.59\uparrow)}$ & $83.24_{(10.01\uparrow)}$ & $69.12_{(16.91\uparrow)}$ & $68.57_{(24.92\uparrow)}$ & $69.18_{(0.25\uparrow)}$ & $57.21_{(4.39\uparrow)}$ \\
 & w/ Intervention & $83.62$ & $83.54$ & $69.12$ & $68.95$ & $71.48$ & $62.12$ \\
  \hline
\multirow{3}{*}{Llama2 (7B)} & Direct & $81.83$ & $81.73$ & $77.21$ & $77.00$ & $66.78$ & $52.23$ \\
\cdashline{2-8}
& \methodname & $83.72_{(1.89\uparrow)}$ & $83.54_{(1.81\uparrow)}$ & $83.82_{(6.62\uparrow)}$ & $83.81_{(6.81\uparrow)}$ & $70.68_{(3.90\uparrow)}$ & $59.58_{(7.35\uparrow)}$ \\
 & w/ Intervention & $85.13$ & $85.04$ & $\mathbf{83.82}$ & $\mathbf{83.82}$ & $73.38$ & $65.32$ \\
  \hline
\multirow{3}{*}{Llama2 (13B)} & Direct & $57.53$ & $51.75$ & $77.94$ & $77.10$ & $52.55$ & $52.27$ \\
\cdashline{2-8}
& \methodname & $87.31_{(29.78\uparrow)}$ & $87.29_{(35.53\uparrow)}$ & $79.41_{(1.47\uparrow)}$ & $79.41_{(2.30\uparrow)}$ & $74.48_{(21.93\uparrow)}$ & $66.32_{(14.06\uparrow)}$ \\
 & w/ Intervention & $\mathbf{87.78}$ & $\mathbf{87.71}$ & \colorbox{green!30}{$78.68$} & \colorbox{green!30}{$78.65$} & $\mathbf{75.92}$ & $\mathbf{69.30}$\\
 \hline
\end{tabular}
\end{adjustbox}
\caption{Results on Constraint, PolitiFact, and GossipCop datasets without access to retrieved background information. The best results for each setting are highlighted with bold numbers. The \colorbox{green!30}{number} and the number with $\uparrow$ have the same meaning as in Table. \ref{liarresults}.}
\vspace{-20pt}
\label{mainresultsthree}
\end{table*}

\begin{table*}[t]
\begin{adjustbox}{max width=0.8\linewidth, center}
\begin{tabular}{cc|cc|cc|cc}
 \hline
\multirow{2}{*}{LLMs} & \multirow{2}{*}{Method} & \multicolumn{2}{c}{\textbf{CP}$\longrightarrow$\textbf{G}} & \multicolumn{2}{c}{\textbf{GP}$\longrightarrow$\textbf{C}} & \multicolumn{2}{c}{\textbf{CG}$\longrightarrow$\textbf{P}} \\
\cline{3-8}
 &  & Acc(\%) & Macro-F1(\%) & Acc(\%) & Macro-F1(\%) & Acc(\%) & Macro-F1(\%) \\
\hline

\multirow{2}{*}{FLAN-T5-xl} & Direct & $67.73$ & $52.80$ & $75.32$ & $74.79$ & $55.88$ & $50.72$ \\
 &  \methodname & $68.13_{(0.40\uparrow)}$ & $56.54_{(3.74\uparrow)}$ & $82.40_{(7.0\uparrow)}$ & $82.09_{(7.31\uparrow)}$ & $61.76_{(5.88\uparrow)}$ & $60.92_{(10.19\uparrow)}$ \\
  \hline
\multirow{2}{*}{FLAN-T5-xxl} & Direct & $68.93$ & $52.82$ & $74.80$ & $73.23$ & $52.21$ & $43.65$ \\
 &  \methodname& $69.13_{(0.2\uparrow)}$ & $53.15_{(0.34\uparrow)}$ & $77.44_{(2.64\uparrow)}$ & $76.21_{(2.98\uparrow)}$ & $66.18_{(13.97\uparrow)}$ & $66.17_{(22.52\uparrow)}$ \\
  \hline
\multirow{2}{*}{Llama2 7B} & Direct & $66.78$ & $52.23$ & $81.83$ & $81.73$ & $77.21$ & $77.00$ \\
 &  \methodname & $68.33_{(1.55\uparrow)}$ & $59.33_{(7.10\uparrow)}$ & $81.60_{(-0.24\downarrow)}$ & $81.04_{(-0.69\downarrow)}$ & $83.09_{(5.88\uparrow)}$ & $82.82_{(5.82\uparrow)}$ \\
 \hline
\multirow{2}{*}{Llama2 13B} & Direct & $52.55$ & $52.27$ & $57.53$ & $51.75$ & $77.94$ & $77.10$ \\
 & \methodname & $70.93_{(18.38\uparrow)}$ & $60.90_{(8.63\uparrow)}$ & $85.09_{(27.56\uparrow)}$ & $84.87_{(33.1\uparrow)}$ & $79.41_{(1.47\uparrow)}$ & $79.41_{(2.30\uparrow)}$\\
 \hline
\end{tabular}
\end{adjustbox}
\caption{Results on cross-domain experiments. \textbf{C}, \textbf{P} and \textbf{G} represent Constraint, PolitiFact, and GossipCop datasets.}
\vspace{-15pt}
\label{crossdomainresults}
\end{table*}

\subsection{Explainability Verification} 
Explainability is a fundamental factor for establishing trust in AI technology. We demonstrate that our framework satisfies this aspect through its inherent mechanism and the visualization of rules.

Unlike approaches that rely heavily on LLMs, our cognition system incorporates expert knowledge to construct a more well-grounded worldview by generating well-defined question templates and logic predicates. Moreover, our decision system can learn interpretable rules from data to deduce logic clauses to debunk fake news by converging implicit parameters to conjunctive and disjunctive semantics. These symbolic units (e.g., questions and logic atoms) and the interpretable DNF Layer contribute to our framework's overall explainability and transparency.

However, as the number of conjunctive and disjunctive layers grows,  it is difficult for human beings to investigate logic rules derived from our decision system. To address this issue, we propose a strategy to prune unnecessary weights in the DNF Layer. 
For example, we present the rules extracted from the pruned model for GossipCop in Table \ref{rulesgossip}, where each conjunctive clause identifies one candidate rule.  The pruning algorithm and rules for other datasets are described in Appendix \ref{sec:appendixc}.

Table \ref{rulesgossip} can be interpreted as learning DNF rules for both true and false labels of input news. Specifically, the true label is predicted if either $\neg \mathrm{conj}_{34}$ or $\neg \mathrm{conj}_{43}$ is true, i.e., either $\neg \mathrm{P}_2 \land \mathrm{P}_3 \land \mathrm{P}_6 \land \mathrm{P}_8$ or $\mathrm{P}_3 \land \mathrm{P}_6 \land \mathrm{P}_8$ is false when removing the negation. Given the semantics of these logic predicates shown in Table \ref{tab:logic-table}, we know that $\mathrm{P}_2$, $\mathrm{P}_3$ and $\mathrm{P}_8$ check the consistency between the background information and a given message, whereas $\mathrm{P}_6$ scrutinizes improper intention from the message alone. On the other hand, the news will be predicted as false if $\mathrm{conj}_{27}$ is true, i.e., $\mathrm{P}_4$ is false which means that the background information in the message is neither accurate or objective according to Table \ref{tab:logic-table}.

\begin{table}[h]
\begin{adjustbox}{max width=0.62\linewidth, center}
\begin{tabular}{l}
\hline
$\mathrm{conj}_{34} = \neg \mathrm{P}_2 \land \mathrm{P}_3 \land \mathrm{P}_6 \land \mathrm{P}_8$  \\
$\mathrm{conj}_{43} = \mathrm{P}_3 \land \mathrm{P}_6 \land \mathrm{P}_8$  \\
$\mathrm{conj}_{27} = \neg \mathrm{P}_4$  \\
$\mathrm{P}_{\mathrm{true}} = \neg \mathrm{conj}_{34} \lor \neg \mathrm{conj}_{43}$ \\
$\mathrm{P}_{\mathrm{false}} = \mathrm{conj}_{27} $\\
\hline
\end{tabular}
\end{adjustbox}
\caption{Extracted rules for the GossipCop dataset when using Llama2 (13B)}
\vspace{-20pt}
\label{rulesgossip}
\end{table}

\subsection{Generalizability Verification}
Ensuring the generalization ability of fake news decision systems is vital for their sustainable and practical deployment. As observed in Table~\ref{crossdomainresults}, {\methodname} consistently outperforms \textit{Direct} across all domains and LLMs without the assistance of any generalization algorithm, while only exhibiting a negligible performance drop in the \textbf{GP}$\longrightarrow$\textbf{C} domain using Llama2 7B. This is attributed to the remarkable zero-shot ability of LLMs and the effectiveness of the DNF layer which further compensates for biased predictions made by LLMs through rule-based aggregation. Particularly, the performance gains of {\methodname} in cross-domain and in-domain experiments (refer to Table~\ref{mainresultsthree}) are positively correlated, implying that the decision system manages to learn domain-agnostic rules. Moreover, the Pearson correlation coefficient between these two groups of performance gains shows a substantial improvement from 0.01 to 0.53 when transitioning from the FLAN-T5 series to the more powerful Llama2 series. This finding suggests that leveraging stronger LLMs to drive the cognition system enhances the generalization capability of our framework.

\subsection{Controllability Verification}
Controllability ensures that fake news detection systems are subject to effective human oversight and intervention. We demonstrate  {\methodname} satisfies this attribute from two aspects. Firstly, we verify the feasibility of manually rectifying rules learned by our decision system that may exhibit irrational behavior. For instance, we observe that $\mathrm{P}_3$ (i.e., "The message contains adequate background information") should have a positive logical relation with $\mathrm{P}_{true}$ instead of negation in  Table \ref{rulesgossip}. To correct this, we perform a manual adjustment by setting the corresponding weight to zero, effectively removing $\mathrm{P}_3$ from the logic rule. However, this modification only leads to a negligible improvement in the test split. Further investigation reveals that the truth value of logic atoms pertaining to $\mathrm{P}_3$ of most real samples is negative, possibly due to the preference of LLMs. This suggests the superiority of our logic-based decision system in reducing the negative effect of incorrect predictions made by LLMs automatically. 
Secondly, we simulate human experts by intervening in the actions of our cognition system. We achieve this by guiding LLMs to expand the question template set $\mathcal{Q}$ using Algorithm \ref{algo:cand_question_generation}, referred to as \textit{w/ intervention} in Tables \ref{liarresults} and \ref{mainresultsthree}. The new question template set for intervention is shown in Table \ref{tab:logic-evo-table}. The results consistently indicate that \textit{w/ intervention} outperforms {\methodname}, highlighting the potential of LLMs as an agency for automatically regulating the behaviors of the cognition system. Thus, our framework ensures a comprehensive control mechanism by simultaneously facilitating human and AI agents' oversight.

Furthermore, we conduct additional experiments to verify the effectiveness of the DNF Layer in logic formulation over other decision systems, namely decision trees, Naive Bayes classifiers and MLP. We replace the DNF Layers with these three algorithms to derive the final decisions. The results are shown in Tables \ref{comparisionamongthreedmodels} and \ref{crossdomainresultsdecision} for in-domain and cross-domain settings, respectively in Appendix \ref{sec:appendixd}.

\section{Conclusion}
In this work, we address the limitations of existing fake news detection methods, which struggle to establish reliability and end-user trust. To tackle this issue, we identify three crucial aspects for constructing trustworthy misinformation detection systems: explainability, generalizability, and controllability. By prioritizing these principles, we propose a dual-system framework {\methodname} that incorporates cognition and decision systems. To validate our framework's feasibility, explainability, generalizability, and controllability, we conduct extensive experiments on diverse datasets and LLMs. These results affirm the effectiveness and trustworthiness of our approach and highlight its significant potential through evolving both subsystems in the future. While we achieve trustworthiness from an algorithmic perspective, we emphasize the importance of further research to improve the trustworthiness of the entire lifecycle of fake news detection systems.

\section*{Limitations}
We identify three main limitations of our work. Firstly, although our framework focuses on enhancing the trustworthiness of fake news detection algorithms, trustworthiness is also influenced by other stages of the AI system lifecycle, such as data collection and deployment. Given the advancements in AI techniques and the importance of online information security, we encourage future research to address the challenges of building trustworthy AI systems comprehensively.

Secondly, as shown in Table \ref{liarresults}, integrating external tools to acquire high-quality background knowledge significantly improves the performance of fake news detection systems. However, collecting information that can effectively support detection tasks using such tools is non-trivial due to the complexities of open-domain information retrieval and the diversity of news content. For instance, we search for background information by inputting check-worthy claims of $\mathrm{P}_1$ into a search engine and filter out as much useful information as possible using GPT-3.5-turbo. However, integrating this evidence led to a slight performance drop on Constraint, PolitiFact, and GossipCop datasets (Due to page limitations, we do not include this experiment in our paper). Therefore, we leave this for future research.

Thirdly, despite the excellent and robust performance of our decision system, especially in generalization ability, the expressiveness of the DNF Layer is still limited due to its simple architecture. For example, the DNF Layer learns rules from data without considering the semantics of logic predicates. It may be crucial to develop more powerful decision models to fully unleash the potential of large language models, such as incorporating the semantics of logic predicates. However, given the low-dimensional input and the need for trustworthiness, the DNF layer remains a prudent choice. Moreover, there also exists a trade-off between trustworthiness and the complexity of the decision system. 

\section*{Ethics Statement}
This paper adheres to the ACM Code of Ethics and Professional Conduct. Specifically, the datasets we utilize do not include sensitive private information and do not pose any harm to society. Furthermore, we will release our codes following the licenses of any utilized artifacts. 

Of paramount importance, our proposed dual-system framework serves as an effective measure to combat fake news and safeguard individuals, particularly in the current era dominated by large generative models that facilitate the generation of deceptive content with increasing ease. Moreover, our approach fulfills explainability, generalizability, and controllability, thereby mitigating concerns regarding the security of AI products and enabling their deployment in real-world scenarios.

\section*{Acknowledgment}
This research is supported by the Ministry of Education, Singapore, under its Academic Research Fund Tier 1 (023618-00001, RG99/23), Research Matching Grant Scheme, Hong Kong Government (9229106). We thank the SAC, PC and Editors of ACL 24/ARR February to read our appealing letter and coordinate the review.  

\newpage
\bibliographystyle{acl_natbib}
\bibliography{ref.bib}

\clearpage

\appendix


\clearpage
\section{Details of Cognition System}
\label{sec:appendix}
Unlike convolutional deep learning-based fake news detection frameworks that classify in a latent space, the cognition system of {\methodname}, aims to emulate human fact-checking experts by complying with specific policies to ensure transparency and controllability of the detection process. In this section, we describe the construction of the set of question templates $\mathcal{Q}$ and $\mathcal{Q}'$ for {\methodname} and $w/ Intervention$ respectively in Appendix \ref{sec:appendixqt}. Furthermore, we introduce a trick for batch training by fixing the number of logic atoms for different inputs in Appendix \ref{sec:appendixtrick} and outline some potential techniques for further improvement of the cognition system in Appendix \ref{sec:appendixpoten}.
\raggedbottom

\subsection{Construction of Question Templates}
\label{sec:appendixqt}
To provide an overview, we present the referenced human-checking process in Table~\ref{tab:fact-check-process}. In this table, Steps \uppercase\expandafter{\romannumeral1}, \uppercase\expandafter{\romannumeral6} and \uppercase\expandafter{\romannumeral7} are excluded from detection algorithms, as they either fall into the preliminary procedures or the post-processing stages of the fake news detection pipeline. These steps may involve data crawling, human-computer interaction, machine translation, etc. As a result, we concentrate on the other steps.

Subsequently, we decompose the process into a Yes/No question template set $\mathcal{Q}$, where each template $\mathrm{Q}_i$ in $\mathcal{Q}$ corresponds to a predicate $\mathrm{P}_i$ in the predicate set $\mathcal{P}$. All question templates and their corresponding predicates are listed in Table~\ref{tab:logic-table}. Specifically, for $\mathrm{Q}_1$, our objective is to determine the trustworthiness of statements in the input news. Here, statements represent crucial information in news articles, playing a vital role in debunking misinformation. Additionally, extracting statements from news is a challenging task. While previous studies like \citet{evikdd23, infosurgeon} used pre-trained language models to generate summaries as statements, we choose to utilize GPT-3.5-turbo to generate statements for simplicity in implementation. The prompt used for this purpose is as follows:
\begin{shaded*}
{\noindent To verify the MESSAGE, what are the critical claims related to this message we need to verify? Please use the following format to answer. If there are no important claims, answer “not applicable”. \\\\
MESSAGE: \\
CLAIM: \\
CLAIM: \\\\
MESSAGE: \$MESSAGE\$.}
\end{shaded*}
Then, we replace the "\$MESSAGE\$" with input news and take the generated claims as statements for $\mathrm{Q}_1$ ($\mathrm{P}_1$).

Additionally, when verifying the controllability of our framework, we propose adjusting the question template set to deal with the diversity of fake news. While this adjustment should be done by fact-checking experts to ensure the reasonableness of new questions, our empirical findings demonstrate the feasibility of guiding large language models, such as GPT-3.5-turbo, to generate new question templates. These templates are then manually filtered by us to create the final question template set $\mathcal{Q'}$, and the corresponding predicate set $\mathcal{P'}$ for intervention, as outlined in Algorithm \ref{algo:cand_question_generation}. Such human verification is incorporated into our intervention method to ensure more controllability because the main point of controllability is to intervene via human knowledge instead of relying on models entirely. Moreover, such manual checking is not time-consuming, with only a few candidate questions being generated. Table \ref{tab:logic-evo-table} presents these newly added question templates and predicates. The prompt $R$ used in this algorithm is as follows:
\begin{shaded*}
{\noindent Write some questions that can be used to determine whether a news report is misinformation. The questions should be answerable by large language models in a close-book situation without requiring additional information. Please format each question using the <s> and </s> tags, such as <s>A question</s>.}
\end{shaded*}
\subsection{Trick for Batch Training}
\label{sec:appendixtrick}
To enable batch training, we fix the number of logic atoms, denoted as $M_i$ for each predicate  $\mathrm{P}_i$. Specifically, If $M_i<\prod\limits_{{j=1}}^{N_i}|\mathrm{X}_{i,j}|$, we randomly select $M_i$ atoms.  Conversely, if $M_i>\prod\limits_{{j=1}}^{N_i}|\mathrm{X}_{i,j}|$, we pad the vector by $0$ accordingly. In the end, $\bm{\mu}$ can be represented as $[\mu_{1,1},\ldots,\mu_{1,M_1},\ldots, \mu_{N,1},\ldots,\mu_{N,M_N}]$, where $\bm{\mu} \in \mathbb{R}^M$ and $M=\sum\limits_i^N M_i$. 

\subsection{The Potential of Cognition System}
\label{sec:appendixpoten}
It is noteworthy that specific techniques can be employed to improve the performance of our cognitive system. For instance, when obtaining the answers to questions as truth values for corresponding logic atoms in Sec.~\ref{sec:logicevaluationllm}, we exclusively consider "Yes" and "No" tokens. However, considering the relationship between model outputs and final predictions, "Right" and "Wrong" tokens can also be suitable candidates. Therefore, drawing motivation from \citep{verberlizer1, verberlizer2}, existing manual or automatic verbalizer techniques that establish mappings between diverse model outputs and final labels can be leveraged to enhance performance. Additionally, the ensemble of prompts, similar to "Yes or No? The answer is: ", has proven effective for the "Yes" and "No" classification task in \citep{yesnoq}. Consequently, our dual-system framework exhibits substantial potential for future improvements in the cognitive system.

\begin{algorithm}[!htbp]
\caption{Question Template Generation for Intervention Algorithm}
\label{algo:cand_question_generation}
\begin{algorithmic}[1]
\REQUIRE{Prompt $R$, the original question template set $\mathcal{Q}$, and a copy of $\mathcal{Q}$ denoted as $\hat{\mathcal{Q}}$}
\ENSURE{The question template set $\mathcal{Q}'$ for intervention}
\STATE Set the number of iteration steps as $T$
\FOR{$\text{Iteration } t = 1, \ldots, T$}
    \STATE Use $R$ to guide GPT-3.5-turbo in generating a  set of new question templates $\mathcal{Q}'$
    \FOR{each question template $\mathrm{Q}'_i$ in $\mathcal{Q}'$}
        \STATE Compute the average similarity score between $\mathrm{Q}'_i$ and all templates in $\hat{\mathcal{Q}}$ using Sentence BERT.
    \ENDFOR
    \STATE Add $\mathrm{Q}'_i \in \mathcal{Q}'$ with the lowest similarity score to $\hat{\mathcal{Q}}$.
\ENDFOR
\STATE $\mathcal{Q}' = \hat{\mathcal{Q}} \setminus \mathcal{Q}$
\STATE Manually refine $\mathcal{Q}'$ by removing duplicate and impractical templates that are non-verifiable through LLMs, resulting in the final $\mathcal{Q}'$.
\end{algorithmic}
\end{algorithm}

\begin{table*}[!htbp]
\begin{adjustbox}{max width=0.8\linewidth, center}
\begin{tabular}{|p\linewidth|}
\hline
\textbf{Step \uppercase\expandafter{\romannumeral1}: Selecting claims} \\
(1) To filter the information on news websites, social media, and online databases through manual selection and computer-assisted selection. \\
(2) The public can submit suspicious claims. \\
(3) Selecting suspicious claims based on their hotness in Hong Kong, considering factors such as the amount of likes, comments, and shares the message has received. \\
\quad A) Is the content checkable? \\
 \quad B) Any misleading or false content? \\
 \quad C) Does it meet public interest? \\
 \quad D) Is it widespread? \\
\hline
\textbf{Step \uppercase\expandafter{\romannumeral2}: Tracing the source} \\
 (1) Determining the source of the information. \\
 (2) Identifying the publication date. \\
 (3) Investigating the publisher and their background and reputation. \\
 (4) Checking for similar information. \\
 (5) Capturing a screen record and attaching the URL link. \\
 (6) Providing two or more additional sources of information. \\
\hline
\textbf{Step \uppercase\expandafter{\romannumeral3}: Fact-checking the suspicious information}  \\
 (1) Applying the Five Ws and an H: When, Where, Who, What, Why, How. \\
 (2) Searching for evidence to verify the information, such as official press releases, authoritative media reports, and research reports. \\
 (3) Attempting to engage the person or organization making the claim through email or telephone, if necessary. \\
 (4) Consulting experts in the relevant field, if necessary. \\
\hline
\textbf{Step \uppercase\expandafter{\romannumeral4}: Retrieving contextual information} \\
 (1) Checking if the original claim contains adequate background information. \\
 (2) Assessing the accuracy and objectivity of the background information. \\
 (3) Identifying any intentionally eliminated content that distorts the meaning. \\
\hline
\textbf{Step \uppercase\expandafter{\romannumeral5}: Evaluating improper intentions} \\
 (1) Assessing if there is any improper intention (e.g., political motive, commercial purpose) in the information. \\
 (2) Investigating if the publisher has a history of publishing information with improper intentions. \\
\hline
\textbf{Step \uppercase\expandafter{\romannumeral6}: Self-checking} \\
 (1) Fact-checkers signing a Declaration of Interest Form before joining the team. \\
 (2) Ensuring fact-checkers maintain objectivity and avoid biases during the process. \\
 (3) Upholding the principle of objectivity and avoiding emotional involvement. \\
\hline
\textbf{Step \uppercase\expandafter{\romannumeral7}: Publishing and reviewing reports} \\
 (1) Completing a draft of the fact-check report, followed by editing and reviewing by professional editors and consultants. \\
 (2) Updating the report if any mistakes or defects are found, and providing clarification on correction reasons and date. \\
\hline
\end{tabular}
\end{adjustbox}
\caption{Fake news detection policy of HKBU FACT CHECK Team \citep{HKBU}}
\label{tab:fact-check-process}
\end{table*}

\begin{table*}[!htbp]
\begin{adjustbox}{max width=0.8\linewidth, center}
\begin{tabular}{|p{0.38\linewidth}|p{0.32\linewidth}|p{0.33\linewidth}|}
\hline
\multicolumn{1}{|c|}{\textbf{Question Template}} & \multicolumn{1}{c|}{\textbf{Logic Predicate: Logic Semantics}} & \multicolumn{1}{c|}{\textbf{Annotation}} \\
\hline
$\mathrm{Q}_1$: Background Information: $\mathrm{X}_{1,1}$. Statement: $\mathrm{X}_{1,2}$. Is the statement true? & $\mathrm{P}_1(\mathrm{X}_{1,1}, \mathrm{X}_{1,2})$: Given the background information $\mathrm{X}_{1,1}$, the statement is true. &$\mathrm{X}_{1,1}$: Background information for input news, $\mathrm{X}_{1,2}$: Check-worthy statements in input news. \\
\hline
$\mathrm{Q}_2$: Background Information: $\mathrm{X}_{2,1}$. Message: $\mathrm{X}_{2,2}$. Is the message true? & $\mathrm{P}_2(\mathrm{X}_{2,1}, \mathrm{X}_{2,2})$: Given the background information $\mathrm{X}_{2,1}$, the message is true. &$\mathrm{X}_{2,1}$: Background information for input news, $\mathrm{X}_{2,2}$: Input news. \\
\hline
$\mathrm{Q}_3$: Message: $\mathrm{X}_{3,1}$. Did the message contain adequate background information? &$\mathrm{P}_3(\mathrm{X}_{3,1})$: The message contains adequate background information. & $\mathrm{X}_{3,1}$: Input news. \\
\hline
$\mathrm{Q}_4$: Message: $\mathrm{X}_{4,1}$. Is the background information in the message accurate and objective? &$\mathrm{P}_4(\mathrm{X}_{4,1})$: The background information in the message is accurate and objective. &$X_{4,1}$: Input news.\\
\hline
$\mathrm{Q}_5$: Message: $\mathrm{X}_{5,1}$. Is there any content in the message that has been intentionally eliminated with the meaning being distorted? &$\mathrm{P}_5(\mathrm{X}_{5,1})$: The content in the message has been intentionally eliminated with the meaning being distorted &$X_{5,1}$: Input news. \\
\hline
$\mathrm{Q}_6$: Message: $\mathrm{X}_{6,1}$. Is there an improper intention (political motive, commercial purpose, etc.) in the message? & $\mathrm{P}_6(\mathrm{X}_{6,1})$: The message has an improper intention. & $X_{6,1}$: Input news. \\
\hline
$\mathrm{Q}_7$: Publisher Reputation: $\mathrm{X}_{7,1}$. Does the publisher have a history of publishing information with an improper intention? &$\mathrm{P}_7(\mathrm{X}_{7,1})$: Given the  publisher reputation $\mathrm{X}_{7,1}$, the publisher has a history of publishing information with an improper intention. &$X_{7,1}$: Publishing history. \\
\hline
$\mathrm{Q}_8$: Background Information: $\mathrm{X}_{8,1}$. Message: $\mathrm{X}_{8,2}$. Is the message false? & $\mathrm{P}_8(\mathrm{X}_{8,1}, \mathrm{X}_{8,2})$: Given the background information $\mathrm{X}_{8,1}$, the message is false. &$\mathrm{X}_{8,1}$: Background information for input news, $\mathrm{X}_{8,2}$: Input news. \\
\hline
\end{tabular}
\end{adjustbox}
\caption{Question template set $\mathcal{Q}$ and logic predicate set $\mathcal{P}$}
\label{tab:logic-table}
\end{table*}

\begin{table*}[!htbp]
\begin{adjustbox}{max width=0.8\linewidth, center}
\begin{tabular}{|p{0.4\linewidth}|p{0.4\linewidth}|p{0.25\linewidth}|}
\hline
\multicolumn{1}{|c|}{\textbf{Question Template}} & \multicolumn{1}{c|}{\textbf{Logic Predicate: Logic Semantics}} & \multicolumn{1}{c|}{\textbf{Annotation}} \\
\hline
$\mathrm{Q}_9$: News Report: $\mathrm{X}_{9,1}$. Is the news report based on facts or does it primarily rely on speculation or opinion? & $\mathrm{P}_9(\mathrm{X}_{9,1})$: The news report is based on facts and relies on speculation or opinion. &$\mathrm{X}_{9,1}$: Input news.\\
\hline
$\mathrm{Q}_{10}$: News Report $\mathrm{X}_{10,1}$: Are there any logical fallacies or misleading arguments present in the news report? & $\mathrm{P}_{10}(\mathrm{X}_{10,1})$: The news report has logical fallacies or misleading arguments. &$\mathrm{X}_{10,1}$: Input news. \\
\hline
$\mathrm{Q}_{11}$: Message: $\mathrm{X}_{11,1}$. Does the message exhibit bias? &$\mathrm{P}_{11}(\mathrm{X}_{11,1})$: The message exhibits bias. & $\mathrm{X}_{11,1}$: Input news. \\
\hline
$\mathrm{Q}_{12}$: News report: $\mathrm{X}_{12,1}$. Are there any grammatical or spelling errors in the news report that may indicate a lack of professional editing?? &$\mathrm{P}_{12}(\mathrm{X}_{12,1})$: The news report has grammatical and spelling errors. &$X_{12,1}$: Input news.\\
\hline
$\mathrm{Q}_{13}$: News report: $\mathrm{X}_{13,1}$. Does the news report use inflammatory language or make personal attacks? &$\mathrm{P}_{13}(\mathrm{X}_{13,1})$: The news report uses inflammatory language and makes personal attacks. &$X_{13,1}$: Input news. \\
\hline
\end{tabular}
\end{adjustbox}
\caption{Question template set $\mathcal{Q}'$ and logic predicate set  $\mathcal{P}'$ generated by GPT-3.5-turbo for intervention}
\label{tab:logic-evo-table}
\end{table*}

\clearpage
\section{Details of Experimental Setting}
\label{sec:appendixb}
\subsection{Datasets}
\noindent\textbf{LIAR} is a publicly available dataset for fake news detection, sourced from POLITIFACT.COM. This dataset comprises six fine-grained labels for truthfulness ratings: $\mathrm{true}$, $\mathrm{mostly true}$, $\mathrm{half true}$, $\mathrm{barely true}$, $\mathrm{false}$, and $\mathrm{pants fire}$. To align with the binary classification problem, we merge $\mathrm{true}$, $\mathrm{mostly true}$ into $\mathrm{true}$ and merge  $\mathrm{barely true}$, $\mathrm{false}$, and $\mathrm{pants fire}$ into $\mathrm{false}$, following \citep{evikdd23}. Moreover, \citet{Liar, liarwithevidence} curated relevant evidence from fact-checking experts (e.g., publisher information, background information, etc.), which serve as gold knowledge in an open setting.

\noindent\textbf{Constraint} is a manually annotated dataset of real and fake news related to COVID-19. We adopt the data pre-processing procedures described in \citep{Constraint}, which involve removing all links, non-alphanumeric characters, and English stop words.

\noindent\textbf{PolitiFact} and \textbf{GossipCop} are two binary classification subsets extracted from FakeNewsNet \citep{Fakenewsnet}. The PolitiFact subset comprises political news, while the GossipCop subset comprises entertainment stories. To optimize experimental costs and adhere to maximum context limitations, we exclude news samples longer than 3,000 words.

For dataset partitioning, we follow the default partition if specified; otherwise, we use a 7:1:2 ratio. Table \ref{datasets} presents the statistics of each dataset.
\begin{table}[!htbp]
  \centering
  \begin{adjustbox}{max width=1.0\columnwidth}
  \begin{tabular}{c|cccc}
    \hline
Split&LIAR&Constraint&PolitiFact&GossipCop\\
    \hline
    Train &10202&6299&469&6999\\
    Validation&1284&2139&66&999\\
    Test&1271&2119&136&2002\\
    \hline
\end{tabular}
\end{adjustbox}
\caption{Statistics of four benchmarks}
\label{datasets}
\end{table}

\subsection{Data Leakage Analysis}
In our work, we used four publicly available datasets to evaluate our proposed framework, {\methodname}. To begin with, following recent work \citep{iclr24dataleakage}, we refer to the problem of data leakage (data contamination) as the situation where the pretraining and finetuning dataset of LLMs contains the testing splits of datasets used in our work. To mitigate the risks associated with data leakage during our evaluation, we took three precautionary steps to ascertain that the probability of the occurrence of data leakage is particularly low:

\noindent\textbf{Manual Check}: For the open-public Flan-T5 and Llama 2 series, we double-checked the dataset cards of these two model families and did not find a data leaking problem. Concretely, we checked the finetuning data  (i.e., Appendix F Finetuning Data Card of \citep{flant5}) and pre-training data (i.e., C4 dataset in Sec. 3.4.1 of \citep{C4dataset}) for the family of Flan-T5 models and checked the pre-training data of Llama 1 (i.e., Sec. 2.1 of \citep{llama1})  while the pre-training data of Llama 2 seems not publicly available yet.

\noindent\textbf{Assumption Experiment}: If data leakage were present, we would expect the detection accuracy of LLMs to scale with model size, given that the memorization ability of LLMs is positively correlated to the size of models empirically \citep{scalinglaw}. However, our results in Tables \ref{liarresults} and \ref{mainresultsthree} do not support this hypothesis, suggesting a low likelihood of data leakage.

\noindent\textbf{Empirical Analysis}: Some measurements for data leakage exist \citep{iclr24dataleakage, llama2}. We used the Sharded Rank Comparison Test, proposed by  \citet{iclr24dataleakage} to analyze potential data leakage in our datasets on Llama2 (7B). We did not analyze the data leakage problem of the GPT series here due to the limited and expensive access, while Llama2 and FLAN-T5 are LLMs we mainly use. The results in Table \ref{testshared} indicate no data leakage risk for Llama2 (i.e., when the p-value> 0.05 means there is no data leakage risk). However, these measurements of data leakage problems may compromise the accuracy of determining whether dataset contamination occurs and have contributed to evaluation performance sometimes because of many confounding factors (a detailed discussion in A.6 of \citep{llama2}).

While {\methodname} has shown satisfactory accuracy on four open-public datasets, our main contribution is the systematic framework that adheres to explainability, generalizability, and controllability. As per our experimental results, {\methodname}’s detection performance can scale by integrating more powerful LLMs and external techniques, demonstrating the effectiveness of our approach as LLMs and related techniques continue to evolve. Consequently, even if the possible data leakage problem may have a deceptively good influence on the detection accuracy, we argue that it will not decrease our work’s contribution.

\begin{table}[!htbp]
\centering
\caption{The Sharded Rank Comparison Test for data leakage problem. We run this test on all testing splits of four datasets for Llama2 (7B).
}
\begin{adjustbox}{max width=1.0\columnwidth}
\begin{tabular}{c|c}
\hline
Dataset    & P-value \\
\hline
LIAR       & 0.8355 \\
Constraint & 0.7869 \\
PolitiFact & 0.7712 \\
GossipCop  & 0.7802 \\
\hline
\end{tabular}
\end{adjustbox}
\label{testshared}
\end{table}

\subsection{Illustration of Different Baselines} 
We compare our model against \textit{Direct}, \textit{Few-shot Direct}, \textit{Zero-shot COT},  \textit{Few-shot COT}, \textit{Few-shot Logic}. \textit{Direct} utilizes LLMs to calculate the probability of each label using Eqs.~\ref{eq1}-\ref{eq2} and then selects the label with the highest likelihood as the predicted label. Building upon \textit{Direct}, \textit{Few-shot Direct} incorporates demonstration samples with known labels as contextual information to enhance the model's performance. \textit{Zero-shot COT} and \textit{Few-shot COT} employ the chain-of-thought (COT) technique \citep{wei2022chain}, enabling LLMs to engage in step-by-step reasoning. While \textit{Zero-shot COT} immediately adds the prompt "Let us think step by step!", \textit{Few-shot COT} provides multiple COT exemplars. For \textit{Few-shot Logic}, we replace the thought process in COT with instantiated questions accompanied by corresponding answers generated by our cognition system. We omit comparisons with Few-shot and COT-based prompt methods for Llama 2 and FLAN-T5 because COT prompts have been found to yield performance gains basically when used with models of approximately 100B parameters \citep{wei2022chain}, and both Few-shot and COT-based methods show no additional improvement over \textit{Direct} as revealed by Table \ref{differentprompts}, we exclusively implement COT-related methods using GPT-3.5-turbo. 

Below we show the templates for these five baselines for the fake news detection task in the closed setting without access to any external knowledge source.

\noindent\textit{\textbf{Direct}}:
\begin{shaded*}
{\noindent Message: \$MESSAGE\$. \\
Is the message  \$Label\$? \\
Yes or No? Response: }
\end{shaded*}
Then, we replace the "\$MESSAGE\$" with input news, "\$Label\$" with candidate truthfulness labels.

\noindent\textit{\textbf{Few-shot Direct}}:
\begin{shaded*}
{\noindent Following given examples to answer Yes/No questions.\\\\
Message: Says the Annies List political group supports third-trimester abortions on demand. \\
Is the message true? \\
Yes or No? Response: No\\\\
Message: Says the Annies List political group supports third-trimester abortions on demand. \\
Is the message false? \\
Yes or No? Response: Yes \\\\
(· · · more examples here · · ·)\\\\
Message: \$MESSAGE\$. \\
Is the message  \$Label\$? \\
Yes or No? Response: }
\end{shaded*}
Then, we replace the "\$MESSAGE\$" with input news, "\$Label\$" with candidate truthfulness labels. Furthermore, during the testing phase, the examples are randomly selected from the training set.
\raggedbottom

\noindent\textit{\textbf{Zero-shot COT}}:
\begin{shaded*}
{\noindent You will be provided with a statement, and your task is to classify its truthfulness into one of two categories: true and false.\\
Message: \$MESSAGE\$. \\
Let's think step by step and give answer with the suffix ``So the final answer is".
}
\end{shaded*}
Then, we replace the "\$MESSAGE\$" with the input news.

\noindent\textit{\textbf{Few-shot COT}}:
\begin{shaded*}
{\noindent You will be provided with a statement, and your task is to classify its truthfulness into one of two categories: true and false.\\\\
Example One\\
Message: Says the Annies List political group supports third-trimester abortions on demand.  \\
Let's think step by step and give answer with suffix ``So the final answer is". \\
Annie's List was comfortable with candidates who oppose more limits on late-term abortions while he also supported candidates who voted for more limits this year.  Both dose not mention of third-trimester abortions. \\
So the final answer is false.\\\\
(· · · more examples here · · ·)\\\\
Message: \$MESSAGE\$. \\
Let's think step by step and give answer with the suffix ``So the final answer is".
}
\end{shaded*}

Then, we replace the "\$MESSAGE\$" with the input news.
\raggedbottom

\noindent\textit{\textbf{Few-shot Logic}}:
\begin{shaded*}
{\noindent You will be provided with a statement, and
 your task is to classify its truthfulness into one of two categories: true and false.\\\\
Example One\\
Message: Says the Annies List political group supports third-trimester abortions on demand.  \\
Decomposed Questions: \\
(1) Statement: The Annies List is a political group. Is the statement true?\\
Yes \\
(2) Statement: The Annies List supports third-trimester abortions. Is the statement true?\\
No\\
(3) Did the message contain adequate background information?\\
False \\\\
(· · · more examples here · · ·)\\\\
Message: \$MESSAGE\$. \\
Let's think step by step and give answer with the suffix ``So the final answer is".
}
\end{shaded*}
Then, we replace the "\$MESSAGE\$" with the input news.

Additionally, we conducted supplementary experiments comparing our framework with other non-LLM-based misinformation detectors (referred to as small models following convention), including  BERT\footnote{\url{https://huggingface.co/bert-base-uncased}} and RoBERTa\footnote{\url{https://huggingface.co/FacebookAI/roberta-base}}, presented in Tables \ref{smallmodelsindomain} and \ref{smallmodelscrossdomain} for in-domain and cross-domain settings, respectively. These small models are finetuned on misinformation detection datasets. Especially for the cross-domain setting, we consider each dataset as a separate domain and fine-tune these models using the train split from source domains, choose the model on the validation split of source ones, and report results on the test split from the target domain. Moreover, we do not compare our framework here with existing transfer learning algorithms because we assume the domain label and target domain data are unavailable in our work.
 
\raggedbottom
\subsection{Model Training for Decision System}
In the decision system of our framework, we employ the DNF Layer to learn human-readable rules from data differentially. To train this model, we utilize the Adam optimizer with a learning rate of 1e-3. Regarding the hyperparameters, we search the conjunction number $C$ within the range [10, 20, 30, 40, 50], and the weight decay within the range [1e-3, 5e-4, 1e-4]. Furthermore, to showcase the superiority of our approach, we maintain consistent hyperparameters across different LLMs in each setting. For instance, all hyperparameters of {\methodname} in the closed setting for the binary classification task on the LIAR dataset remain unchanged. The batch size is set to 64, and the number of epochs is set to 30. Additionally, we progressively converge the model towards symbolic semantics by adjusting $\delta$ (refer to Appendix~\ref{sec:appendixe} for detail) to 1 or -1 before the first 15 epochs using exponential decay.

\clearpage

\section{Details of Explainability Study}
\label{sec:appendixc}
To enhance the accessibility of the rules generated by the DNF Layer, we propose a pruning algorithm that extracts more concise logic clauses by eliminating insignificant weights. The algorithm is described in Algorithm \ref{algo:pruningdnf}. Furthermore, to demonstrate the explainability of our framework, we visualize the extracted rules obtained from the pruned model for Constraint, PolitiFact, and GossipCop datasets in Tables \ref{rulesconstraint}, \ref{rulespolitifact} and \ref{rulesgossip}, respectively. In these tables, $\mathrm{P}_{\text{true}}$ and $\mathrm{P}_{\text{false}}$ represent the proposition that the input news is identified as true or false, respectively. In our visualization experiments, we employ Llama2 (13B) as the LLM in the cognition system. We set the number of conjunctive layers $C$ as $50$, the performance drop threshold $\epsilon$ as $0.005$, and $b$ as 0.0001 to reduce the number of conjunction clauses. More details regarding these parameters can be found in Appendix~\ref{sec:appendixe}.

Similar to Symbolic AI, such as expert systems, our learned rules can be intuitively translated into natural language. For instance, consider the rules provided in Table \ref{rulesgossip}, $\mathrm{conj}_{27} = \neg \mathrm{P}_4$, 
$\mathrm{P}_{\mathrm{false}} = \mathrm{conj}_{27} $ and 
the semantics of $\mathrm{P}_4$ in Table 6 is ``The background information in the message is accurate and objective'', $\mathrm{P}_{\mathrm{false}}$  can be translated as ``The input message (news) is \textbf{false} when the background information in the message is \textbf{not} accurate and objective".

\begin{algorithm}[h]
\caption{Pruning Algorithm for the DNF Layer}
\label{algo:pruningdnf}
\begin{algorithmic}[1]
\small
\REQUIRE{Trained DNF Layer $\Phi$, performance drop threshold $\epsilon$}
\ENSURE{Pruned DNF Layer $\Phi'$ and extracted rule set $\mathcal{R}$}
\STATE Initialize $\mathcal{R'}$ as an empty set
\STATE Initialize $\mathcal{R}$ by extracting rules from $\Phi$
\STATE Initialize $\Phi'$ using $\Phi$
\WHILE{$|\mathcal{R'}| \neq |\mathcal{R}|$}
    \STATE Initialize $\mathcal{R}$ by extracting rules from $\Phi'$
    \STATE Prune disjunctions if the removal of a disjunction results in a performance drop smaller than $\epsilon$
    \STATE Prune unused conjunctions that are not utilized by any disjunction
    \STATE Prune conjunctions if the removal of a conjunction results in a performance drop smaller than $\epsilon$
    \STATE Prune disjunctions that use empty conjunctions
    \STATE Prune disjunctions again if the removal of a disjunction results in a performance drop smaller than $\epsilon$
    \STATE Update the pruned model as $\Phi'$ and extract rules from $\Phi'$ to obtain $\mathcal{R'}$;
\ENDWHILE
\end{algorithmic}
\end{algorithm}

\begin{table}[!htbp]
\begin{adjustbox}{max width=\linewidth, center}
\begin{tabular}{l}
\hline
$\mathrm{conj}_{48} = \mathrm{P}_4 \land \neg \mathrm{P}_8$ \\
$\mathrm{conj}_{25} = \neg \mathrm{P}_4 \land \neg \mathrm{P}_5 \land \mathrm{P}_8$  \\
$\mathrm{conj}_{40} = \mathrm{P}_2 \land \mathrm{P}_4$ \\
$\mathrm{P}_{\mathrm{true}} = \mathrm{conj}_{48}$  \\
$\mathrm{P}_{\mathrm{false}} = \mathrm{conj}_{25} \lor \neg \mathrm{conj}_{40}$  \\
\hline
\end{tabular}
\end{adjustbox}
\caption{Extracted rules for the Constraint dataset when using Llama2 (13B).}
\label{rulesconstraint}
\end{table}

\begin{table}[!htbp]
\begin{adjustbox}{max width=\linewidth, center}
\begin{tabular}{l}
\hline
$\mathrm{conj}_{36} = \mathrm{P}_3 \land \mathrm{P}_6 \land \mathrm{P}_8$ \\
$\mathrm{conj}_{44} = \mathrm{P}_5 \land \mathrm{P}_1 \land \mathrm{P}_8$ \\
$\mathrm{conj}_{0} = \mathrm{P}_1$ \\
$\mathrm{conj}_{49} = \mathrm{P}_2\land \mathrm{P}_3 \land \mathrm{P}_4$ \\
$\mathrm{P}_{\mathrm{true}} = \neg \mathrm{conj}_{36} \lor \neg \mathrm{conj}_{44}$ \\
$\mathrm{P}_{\mathrm{false}} = \neg \mathrm{conj}_{0} \lor \neg\mathrm{conj}_{49}$  \\
\hline
\end{tabular}
\end{adjustbox}
\caption{Extracted rules for the PolitiFact dataset when using Llama2 (13B).}
\label{rulespolitifact}
\end{table}

\begin{table*}[!htbp]
\begin{adjustbox}{max width=0.8\linewidth, center}
\begin{tabular}{cc|cc|cc|cc}
 \hline
\multirow{2}{*}{LLMs} & \multirow{2}{*}{Method} & \multicolumn{2}{c}{Constraint} & \multicolumn{2}{c}{PolitiFact} & \multicolumn{2}{c}{GossipCop} \\
\cline{3-8}
 &  & Acc(\%) & Macro-F1(\%) & Acc(\%) & Macro-F1(\%) & Acc(\%) & Macro-F1(\%) \\
 \hline
\multirow{3}{*}{FLAN-T5-xl} & Direct & $75.32$ & $74.79$ & $55.88$ & $50.72$ & $67.73$ & $52.80$ \\
 &  Few-shot& $75.17$ & $74.48$ & $52.20$ & $45.07$ & $67.13$ & $51.20$ \\
 & Few-shot COT & $52.67$ & $45.76$ & $58.08$ & $56.62$ & $46.65$ & $46.50$ \\
  \hline
\multirow{3}{*}{FLAN-T5-xxl} & Direct & $74.80$ & $73.23$ & $52.21$ & $43.65$ & $68.93$ & $52.82$ \\
 & Few-shot & $75.97$ & $75.97$ & $50.73$ & $41.10$ & $68.53$ & $51.87$ \\
 & Few-shot COT & $52.66$ & $45.33$ & $50.61$ & $41.43$ & $65.98$ & $47.15$ \\
  \hline
\multirow{3}{*}{Llama2 (7B)} & Direct & $81.83$ & $81.73$ & $77.21$ & $77.00$ & $66.78$ & $52.23$ \\
\cdashline{2-8}
& Few-shot & $71.68$ & $71.30$ & $75.74$ & $75.74$ & $66.13$ & $59.62$ \\
 & Few-shot COT & $52.10$ & $34.77$ & $55.14$ & $42.89$ & $47.95$ & $47.43$ \\
  \hline
\multirow{3}{*}{Llama2 (13B)} & Direct & $57.53$ & $51.75$ & $77.94$ & $77.10$ & $52.55$ & $52.27$ \\
\cdashline{2-8}
& Few-shot & $57.24$ & $50.48$ & $80.14$ & $79.56$ & $51.55$ & $51.39$ \\
 & Few-shot COT & $53.79$ & $44.98$ & $50.01$ & $33.33$ & $65.28$ & $50.92$\\
 \hline
\end{tabular}
\end{adjustbox}
\caption{Comparison between different prompt methods on FLAN-T5 and Llama2 series.}
\vspace{-15pt}
\label{differentprompts}
\end{table*}

\begin{table*}[!htbp]
\begin{adjustbox}{max width=0.8\linewidth, center}
\begin{tabular}{c|cc|cc|cc|cc}
 \hline
\multirow{2}{*}{Method} & \multicolumn{2}{c}{Constraint} & \multicolumn{2}{c}{PolitiFact} & \multicolumn{2}{c}{GossipCop} &  \multicolumn{2}{c}{LIAR} \\
\cline{2-9}
 & Acc(\%) & Macro-F1(\%) & Acc(\%) & Macro-F1(\%) & Acc(\%) & Macro-F1(\%) & Acc(\%) & Macro-F1(\%) \\
 \hline
 BERT & $96.98$ & $97.11$ & $85.29$ & $85.71$ & $81.97$ & $86.45$ & $63.06$ & $62.42$\\
 RoBERTa & $ \mathbf{97.07}$ & $\mathbf{97.21}$ & $\mathbf{88.97}$ & $\mathbf{89.36}$ & $\mathbf{82.72}$ & $\mathbf{87.07} $ & $64.55$ & $63.16$ \\
 {\methodname} (best)  & $87.78$ & $87.71$ & $83.82$ & $83.82$ & $75.92$ & $69.30$ & $\mathbf{67.73}$ & $\mathbf{66.97}$  \\
 \hline
\end{tabular}
\end{adjustbox}
\caption{Comparison between small models and {\methodname}  on four datasets for \textbf{binary classification task} in an in-domain setting.}
\label{smallmodelsindomain}
\end{table*}

\begin{table*}[!htbp]
\begin{adjustbox}{max width=0.8\linewidth, center}
\begin{tabular}{c|cc|cc|cc}
 \hline
\multirow{2}{*}{Method} & \multicolumn{2}{c}{\textbf{CP}$\longrightarrow$\textbf{G}} & \multicolumn{2}{c}{\textbf{GP}$\longrightarrow$\textbf{C}} & \multicolumn{2}{c}{\textbf{CG}$\longrightarrow$\textbf{P}} \\
\cline{2-7}
 & Acc(\%) & Macro-F1(\%) & Acc(\%) & Macro-F1(\%) & Acc(\%) & Macro-F1(\%) \\
 \hline
 BERT & $46.97$ & $31.12$ & $65.69$ & $70.65$ & $48.53$ & $46.97$ \\
 RoBERTa & $47.45$ & $38.26$ & $64.56$ & $65.79$ & $52.21$ & $48.00$ \\
 {\methodname} (best)  & $\mathbf{70.93}$ & $\mathbf{60.90}$ & $\mathbf{85.09}$ & $\mathbf{84.87}$ & $\mathbf{83.09}$ & $\mathbf{82.82}$ \\
 \hline
\end{tabular}
\end{adjustbox}
\caption{Comparison between small models and {\methodname} for \textbf{ binary classification task} in a cross-domain setting.}
\label{smallmodelscrossdomain}
\end{table*}

\clearpage

\section{Comparison with Different Decision Models}
\label{sec:appendixd}
\begin{table*}[!htbp]
\begin{adjustbox}{max width=0.8\linewidth, center}
\begin{tabular}{cc|cc|cc|cc}
 \hline
\multirow{2}{*}{LLMs} & \multirow{2}{*}{Method} & \multicolumn{2}{c}{Constraint} & \multicolumn{2}{c}{PolitiFact} & \multicolumn{2}{c}{GossipCop} \\
\cline{3-8}
 &  & Acc(\%) & Macro-F1(\%) & Acc(\%) & Macro-F1(\%) & Acc(\%) & Macro-F1(\%) \\
 \hline
\multirow{4}{*}{FLAN-T5-large} & Decision Tree & $78.53$ & $78.30$ & $67.65$ & $67.19$ & $70.88$ & $62.76$ \\
 & Bayes Classifier & $80.93$ & $80.86$ & $66.18$ & $66.15$ & $68.33$ & $61.04$ \\
  & MLP & $81.26$ & $81.16$ & $71.42$ & $63.43$ & $71.62$ & $63.74$ \\
 &  {\methodname}  & $80.32$ & $80.11$ & $67.65$ & $67.65$ & $69.53$ & $59.39$ \\
  \hline
\multirow{4}{*}{FLAN-T5-xl} & Decision Tree & $84.29$ & $84.27$ & $66.91$ & $66.10$ & $71.13$ & $61.58$ \\
 & Bayes Classifier & $82.40$ & $82.22$ & $68.38$ & $67.88$ & $68.23$ & $60.23$ \\
& MLP & $84.52$ & $84.44$ & $70.28$ & $60.74$ & $70.78$ & $ 	62.76$ \\
 &  {\methodname}  & $83.77$ & $83.66$ & $68.82$ & $64.68$ & $69.58$ & $58.72$ \\
  \hline
\multirow{4}{*}{FLAN-T5-xxl} & Decision Tree & $84.14$ & $84.12$ & $72.06$ & $71.00$ & $72.13$ & $67.08$ \\
 & Bayes Classifier & $82.49$ & $82.30$ & $68.38$ & $67.61$ & $68.38$ & $57.62$ \\
 & MLP & $83.29$ & $83.15$ & $72.78$ & $65.82$ & $72.52$ & $65.98$ \\
 &  {\methodname}  & $83.39$ & $83.24$ & $69.12$ & $68.57$ & $69.18$ & $57.21$ \\
  \hline
\multirow{4}{*}{Llama2 (7B)} & Decision Tree & $84.33$ & $84.32$ & $79.41$ & $77.00$ & $72.38$ & $65.24$ \\
 & Bayes Classifier & $83.11$ & $82.97$ & $76.47$ & $76.29$ & $71.98$ & $66.67$ \\
  & MLP & $84.99$ & $84.94$ & $74.68$ & $68.80$ & $74.83$ & $68.86$ \\
&  {\methodname}  & $83.72$ & $83.54$ & $\mathbf{83.82}$ & $\mathbf{83.81}$ & $70.68$ & $59.58$ \\
  \hline
\multirow{4}{*}{Llama2 (13B)} & Decision Tree & $86.50$ & $86.49$ & $83.09$ & $83.07$ & $74.43$ & $68.99$ \\
 & Bayes Classifier & $84.99$ & $84.92$ & $80.15$& $80.06$ & $73.58$ & $69.59$\\
  & MLP & $87.31$ & $87.31$ & $77.37$ & $72.72$ & $\mathbf{76.97}$ & $\mathbf{72.01}$ \\
&  {\methodname}  & $\mathbf{87.31}$ & $\mathbf{87.29}$ & $79.41$ & $79.41$ & $74.48$ & $66.32$ \\
 \hline
\end{tabular}
\end{adjustbox}
\caption{Results of different decision models on Constraint, PolitiFact, and GossipCop datasets without access to retrieved background information. The best results for each dataset are highlighted with bold numbers.}
\label{comparisionamongthreedmodels}
\end{table*}

\begin{table*}[!htbp]
\begin{adjustbox}{max width=0.8\linewidth, center}
\begin{tabular}{cc|cc|cc|cc}
 \hline
\multirow{2}{*}{LLMs} & \multirow{2}{*}{Method} & \multicolumn{2}{c}{\textbf{CP}$\longrightarrow$\textbf{G}} & \multicolumn{2}{c}{\textbf{GP}$\longrightarrow$\textbf{C}} & \multicolumn{2}{c}{\textbf{CG}$\longrightarrow$\textbf{P}} \\
\cline{3-8}
 &  & Acc(\%) & Macro-F1(\%) & Acc(\%) & Macro-F1(\%) & Acc(\%) & Macro-F1(\%) \\
\hline
\multirow{4}{*}{FLAN-T5-xl} & Decision Tree & $68.98$ & $62.33$ & $73.67$ & $73.32$ & $63.97$ & $62.71$ \\
 & Bayes Classifier & $67.13$ & $59.26$ & $82.49$ & $82.49$ & $64.71$ & $64.64$ \\
& MLP & $67.63$ & $55.67$ & $74.80$ & $74.78$ & $64.71$ & $63.76$ \\
 & {\methodname} & $68.13$ & $56.54$ & $82.40$ & $82.09$ & $61.76$ & $60.92$ \\
  \hline
\multirow{4}{*}{FLAN-T5-xxl} & Decision Tree & $68.33$ & $55.53$ & $70.60$ & $70.35$ & $61.03$ & $60.98$ \\
 & Bayes Classifier & $68.33$ & $54.71$ & $82.63$ & $82.51$ & $62.50$ & $62.50$ \\
 & MLP & $67.58$ & $53.96$ & $74.23$ & $74.22$ & $66.18$ & $65.81$ \\
 &  {\methodname}  & $69.13$ & $53.15$ & $77.44$ & $76.21$ & $66.18$ & $66.17$ \\
  \hline
\multirow{4}{*}{Llama2 7B} & Decision Tree & $52.20$ & $52.05$ & $76.40$ & $75.02$ & $66.91$ & $64.84$ \\
 & Bayes Classifier& $65.98$ & $62.46$ & $82.82$ & $82.60$ & $67.65$ & $65.49$ \\
 & MLP & $65.73$ & $64.87$ & $81.50$ & $80.82$ & $75.00$ & $74.65$ \\
 & {\methodname} & $68.33$ & $59.33$ & $81.60$ & $81.04$ & $\mathbf{83.09}$ & $\mathbf{82.82}$ \\
 \hline
\multirow{4}{*}{Llama2 13B} & Decision Tree & $61.59$ & $61.14$ & $71.54$ & $68.21$ & $71.32$ & $71.32$ \\
 & Bayes Classifier & $\mathbf{71.53}$ & $\mathbf{69.09}$ & $82.59$ & $82.25$ & $78.68$ & $78.25$ \\
 & MLP & $71.33$ & $68.48$ & $78.76$ & $77.62$ & $ 	80.15$ & $79.96$ \\
 & {\methodname} & $70.93$ & $60.90$ & $\mathbf{85.09}$ & $\mathbf{84.87}$ & $79.41$ & $79.41$\\
 \hline
\end{tabular}
\end{adjustbox}
\caption{Results of different decision models on cross-domain experiments. \textbf{C}, \textbf{P} and \textbf{G} represent Constraint, PolitiFact, and GossipCop datasets, respectively. The best results for each dataset are highlighted with bold numbers.}
\label{crossdomainresultsdecision}
\end{table*}

In our work, we utilize the DNF Layer to construct our decision system, guaranteeing explainability and controllability. However, there are also other alternatives, such as existing neural symbolic architectures and interpretable machine learning algorithms. By comparing the DNF Layer with these candidates, we demonstrate that our dual-system framework can achieve better performance by inventing a more effective decision model to unleash the ability of LLMs.

While existing neural symbolic architectures can extract useful rules from data \citep{booch2021thinking}, they indeed have certain limitations. Firstly, these architectures often require complex mechanisms to implement logical operations, which makes them unsuitable for immediate application in fake news detection tasks. For example, \citet{qu2020rnnlogic,cheng2022neural} developed neural-symbolic models for knowledge graph completion, but their reliance on well-defined graph structures makes them infeasible for our task. Secondly, these architectures often suffer from efficiency issues. For instance, $\delta$LP proposed in \citep{evans2018learning} had high computational complexity, and HRI \citep{HRI} was incompatible with batch training, which externally required users to pre-define rule templates to constrain the search space. Furthermore, to the best of our knowledge, there may be no neural-symbolic framework available that can simultaneously handle the challenges of missing values and multi-grounding problems (i.e., one predicate can be instantiated as multiple logic atoms), which are common in our tasks. Therefore, we acknowledge the need for future research to develop a more suitable and powerful neural-symbolic framework in the context of fake news detection.

Since each dimension in $\bm{\mu}$ is precisely bonded to a question template (logic predicate), we can employ traditional machine learning classification algorithms, including decision tree\footnote{\url{https://scikit-learn.org/stable/modules/tree.html}}, naive Bayes Classifier\footnote{\url{https://scikit-learn.org/stable/modules/naive_bayes.html}} and multi-layer perceptron (MLP), to replace the DNF Layer to drive our decision system, while maintaining partial aspects of trustworthy AI. Therefore, we compare the DNF Layer with these three methods in both in-domain and cross-domain settings on three datasets, shown in Tables \ref{comparisionamongthreedmodels} and  \ref{crossdomainresultsdecision}, respectively.

According to the results, we conclude that the decision tree and MLP perform better when the training and testing data are from the same domain. Meanwhile, the naive Bayes Classifier demonstrates more satisfactory generalization performance in cross-domain experiments across various LLMs. This implies that our proposed dual-system framework shows potential in developing a more powerful decision module, such as an ensemble of these algorithms. However, the DNF Layer still outperforms these three methods in most cases when using Llama2 (13B) as the driver of the cognition system, achieving a better trade-off between accuracy and generalization ability. Moreover, the DNF Layer also exhibits advantages over these methods in terms of its ability to handle missing values and multi-grounding problems, as well as its flexibility in efficiently searching logic rules in a large space, whereas the decision tree is constrained by depth and width.
\FloatBarrier
\clearpage
\FloatBarrier

\section{Comparison with Existing Work on Three Principles}
\label{sec:appendixf}
It is imperative to compare our LLM-based framework with prevailing misinformation detection methods across dimensions of explainability, generalizability, and controllability. We conduct additional experiments to compare with small models to demonstrate the strength of {\methodname} in generalizability. However, quantitatively measuring explainability and controllability in deep learning is presently challenging \citep{li2023trustworthy}, necessitating substantial research endeavors. 

\noindent\textit{\textbf{Generalizability}}:  
We conduct additional experiments comparing with small models (BERT and RoBERTa) in Tables \ref{smallmodelsindomain} and \ref{smallmodelscrossdomain} for in-domain (with finetuning) and cross-domain settings, respectively. These results illustrate that small models only outperform {\methodname} in an in-domain setting, but {\methodname} excels in zero-shot generalization (around 30\% improvement in terms of Accuracy and F1-Score) and can handle more complex misinformation detection tasks, exemplified by superior performance on the LIAR dataset. This advantage aligns with many real scenarios, characterized by the absence of training data and the presence of sophisticated misinformation \citep{emnlp23towards}. Consequently, {\methodname} proves significantly advantageous in such contexts. 

Moreover, the feasibility and adaptability of {\methodname} are underscored by the resource-intensive nature of gathering adequate data for small models. Additionally, our framework, as a general and systematic framework, can achieve better in-domain accuracy by integrating small fine-tuned models into our cognition system, treating their binary classification outputs as truth values

\noindent\textit{\textbf{Explainability}}:
Current interpretative methods using feature importance, attention visualization, and multiview learning \citep{shu2019defend, eviwww22, evikdd23, YingHZQ0G23} may be unreliable and possess limited explanatory power, as indicated by \citep{LiuLGKL022}. Another approach \citep{liu-etal-2023-interpretable}, employing neural-symbolic learning for multimodal misinformation detection, falls short of clause length and readability caused by its unexplainable predicates. Unlike small-model-based misinformation detectors, our cognition system incorporates expert knowledge to construct a more well-grounded worldview, which is unrealistic for small models to achieve. Furthermore, another group of work  \citep{gptfakenews1, badactor, yue2024evidence, qi2024sniffer} explored large generative language models (e.g., ChatGPT) and regarded the intermediate chain of thoughts as an explanation. Nevertheless, these explanations may not be reliable due to the hallucination phenomenon and the misalignment problem of AGI \citep{chencanyu}. Compared with them, our decision system can learn interpretable rules to explicitly aggregate generated logic atoms for further double-checking instead of relying on the implicit aggregation of LLMs. 

\noindent\textit{\textbf{Controllability}}:  As shown in Sec. 2.2, some studies integrated human-in-loop techniques \citep{humaninloo} for data sampling and model evaluation, whereas our framework prioritizes algorithm design. Moreover, while recent RLHF techniques \citep{dpo} can incorporate human guidance in model behaviors based on reinforcement learning, they indeed require external high-quality fine-tuning data and sophisticated finetuning. In contrast, our framework achieves controllability through natural manipulation of the question set and logic rules in our cognition and decision systems

In summary, {\methodname} effectively addresses challenges in explainability, generalizability, and controllability. We also emphasize {\methodname} is a general framework and does not sacrifice performance for explainability, generalizability, and controllability, considering its potential to integrate fine-tuned small models to improve the in–domain performance. 
\FloatBarrier
\clearpage
\FloatBarrier

\section{Cost Analysis}
\label{sec:appendixg}
One crucial consideration of {\methodname} is the expense associated with the $N$ queries to LLMs. The specific costs, including inference time and token cost, will be discussed below.

\noindent\textit{\textbf{Inference Time}}: Due to the limited access times of GPT-3.5-turbo in minutes, it is time-consuming to perform $N$ queries for our framework. However, it is worthwhile that it may also require multiple queries for GPT-3.5-turbo to adopt self-consistency and least-to-most prompt techniques to achieve the comparable performance as our framework, given there is a performance gap between \textit{Direct} and {\methodname} in Table \ref{liarresults}. 

Furthermore, our experiments indicate that utilizing smaller LLMs, like FLAN-T5 (XL and XXL) and Llama 2 (7B and 13B), suffices for effective misinformation detection. In this case, our framework stands out from COT-based methods \citep{plmprogramming, emnlp23towards, wang2023explainable} as it eliminates the necessity of generating numerous immediate reasoning steps sequentially. Specifically, our cognition system only requires decoding the first token (i.e., "yes"/"no") to compute truth values. Since the primary bottleneck in the inference time of LLMs arises from subsequential decoding, the cost of {\methodname} is lower than COT-based methods. For instance, consider a COT-based model that generates $100$ tokens for input news. The theoretical inference time of our framework is thus $\frac{1}{100}$ of COT-based methods, assuming parallel decoding of the first token of $N$ questions.

\noindent\textit{\textbf{Token Cost}}: Assuming our framework needs $N$ queries and other LLM-based methods requires one with input query length $L$ and output length $M$, $c_{in}$
is the price of input tokens, $c_{out}$ is the price of output tokens, and the token cost ratio between our framework and LLM-based methods is $\frac{N\times (L\times c_{in}+1\times c_{out})}{L\times c_{in}+M \times c_{out}}$. In general, $c_{out}$ is higher than $c_{in}$. Then if $M$ is significantly high when the output of other LLM-based methods contains lots of tokens such as COT, the total cost does not give much difference.

Additionally, we conduct experiments on various language models to verify the versatility of our framework, especially for FLAN-T5-large (780M) in Table 1. That is to say, our framework can build on smaller models (780M) while the size of bert-large has been 334M. While there have been more and more distillation techniques for LLMs to obtain lightweight models, many engineering efforts can be made to reduce the running cost, which is not the focus of our work. Consequently, we conclude that the cost of my framework is acceptable.

\FloatBarrier
\clearpage
\FloatBarrier

\section{Formal Description of DNF Layer}
\label{sec:appendixe}
In this section, we introduce modified Disjunctive Normal Form (DNF) Layer employed in our framework. The DNF Layer is built from semi-symbolic layers (SL), which can progressively converge to symbolic semantics such as conjunction $\land$ and disjunction $\lor$.

Specifically, for the truth value vector $\bm{\mu}\in\mathbb{R}^M$ mentioned in Sec.~\ref{sec:logicevaluationllm}, SL can be formulated as follows:
\begin{align}
\label{eq4}
\mu_o = \mathrm{tanh}\left(\sum_{j}^M w_j\mu_j + \beta\right), \\
\label{eq5}
\beta = \delta\left(b-\sum_j|w_j\mu_j|\right),
\end{align}
where $w_j$ represents learnable parameters, $b=\max\limits_j|w_j\mu_j|$ and $\delta \in [-1, 1]$ represents the semantic gate selector. $\mu_j$ is the truth value for the $j$th logic atom obtained from the cognitive system. The sign of the learned weight $w_j$ indicates whether $\mu_j$ (if $w_j$ is positive) or its negation (if $w_j$ is negative) contributes to $\mu_o$. Thus, logical negation (e.g., $\neg p_j$) can be computed as the multiplicative inverse of the input: $-\mu_j$. 

Eq. \ref{eq4} resembles a standard feed-forward layer, aiming to compute a single truth value from a collection of values $\mu_j$ corresponding to different instantiations of a single predicate/question. $\beta$ serves as the bias term. As shown by \citep{pix2rule}, by adjusting $\delta$ from $0$ to $1$ during training, SL tends to converge to conjunctive semantics as $\mathrm{SL}_\land$ (e.g., $p_1 \wedge p_2, \ldots, \wedge p_M$), indicating that if at least one $w_j\mu_j$ is false, the output $\mu_o$ will be false; otherwise, $\mu_o$ will be true. Conversely, by gradually adjusting $\delta$ from $0$ to $-1$, SL can attain disjunctive semantics as $\mathrm{SL}_\lor$ (e.g., $p_1 \vee p_2, \ldots, \vee p_M$), where if at least one $w_j\mu_j$ is true, $\mu_o$ will be true; otherwise, $\mu_o$ will be false. Additionally, $b$ can guarantee $\mu_o$ being true (false) when all $w_j\mu_j$ are true (false) for $\mathrm{SL}_\land$ ($\mathrm{SL}_\lor$).

Since each dimension in $\bm{\mu}$ corresponds to the same predicate for different inputs, SL effectively represents the relationship among different instantiations and the target output $\mu_o$, enabling the learning of generic rules for various inputs. Moreover, by employing rule-based aggregation, our framework exhibits noise tolerance against incorrect predictions of LLMs in the cognition system, particularly owing to the $\mathrm{SL}_\lor$.  

Notably, one predicate can be instantiated by multiple assignments, i.e., $\mathrm{P}_i$ pertains to $M_i$ logic atoms in Appendix~\ref{sec:appendixtrick}. Thus, the parameters bound to these $M_i$ logic atoms should naturally share the logical semantics of $\mathrm{P}_i$. Instead of gathering all possible combinations of $M_i$ logic atoms for training
($\prod\limits_{j=1}^{M_i}j$),
we let these logic atoms share the same $w$. In this scenario, SL can be represented as follows:

\begin{align}
\label{eq6}
\mu_o &= \mathrm{tanh}(\sum_{i}^N \sum_{j}^{M_i} w_i\mu_{i,j} + \beta), \\
\label{eq7}
\beta &= \delta(b-\sum_{i}^N \sum_{j}^{M_i}|w_i\mu_{i,j}|),
\end{align}
where $N$ is the number of predicates.

\end{document}